\documentclass{article}
\newenvironment{proof-sketch}{\noindent{\bf Sketch of Proof}\hspace*{1em}}{\qed\bigskip}
\newenvironment{proof-idea}{\noindent{\bf Proof Idea}\hspace*{1em}}{\qed\bigskip}
\newenvironment{proof-of-lemma}[1]{\noindent{\bf Proof of Lemma #1}\hspace*{1em}}{\qed\bigskip}
\newenvironment{proof-attempt}{\noindent{\bf Proof Attempt}\hspace*{1em}}{\qed\bigskip}

{\makeatletter
 \gdef\xxxmark{%
   \expandafter\ifx\csname @mpargs\endcsname\relax 
     \expandafter\ifx\csname @captype\endcsname\relax 
       \marginpar{\textcolor{red}{xxx~}}
     \else
       \textcolor{red}{xxx~}
     \fi
   \else
     \textcolor{red}{xxx~}
   \fi}
 \gdef\xxx{\@ifnextchar[\xxx@lab\xxx@nolab}
 \long\gdef\xxx@lab[#1]#2{{\bf [\xxxmark \textcolor{red}{#2} ---{\sc #1}]}}
 \long\gdef\xxx@nolab#1{{\bf [\xxxmark \textcolor{red}{#1}]}}
}

\newcommand{\calA}{\ensuremath{\mathcal{A}}}
\newcommand{\calB}{\ensuremath{\mathcal{B}}}

\newcommand{\calD}{\ensuremath{\mathcal{D}}}

\newcommand{\calL}{\ensuremath{\mathcal{L}}}

\newcommand{\calN}{\ensuremath{\mathcal{N}}}

\usepackage{textcomp, libertine}
\usepackage[utf8]{inputenc} 
\usepackage[T1]{fontenc}    
\usepackage{url}            
\usepackage{booktabs}       
\usepackage{nicefrac}       
\usepackage{microtype}      
\usepackage{multirow}
\usepackage{amsmath,amsfonts,amssymb, amsthm}
\usepackage{algorithm, algorithmic}
\usepackage{tikz}
\usepackage{pgfplots}
\usepgfplotslibrary{colorbrewer}
\usepackage{mathtools}
\usepackage{fixmath}
\usepackage{arydshln}
\usepackage{wrapfig}
\usepackage{enumitem}
\usepackage{graphicx}
\usepackage{transparent}
\usepackage{tabularx}

\usepackage{float}

\usepackage{epstopdf}
\usepackage{caption}
\usepackage{subcaption}
\usepackage{xcolor}
\usepackage{caption}
\usepackage{calc}
\usepackage{relsize}

\pgfplotsset{compat=1.10}
\usepgfplotslibrary{statistics}
\usetikzlibrary{decorations.pathreplacing} 

\usepackage{import}
\usepackage{pdfpages}

 \newcommand{\miniscule}{\@setfontsize\miniscule{4}{3}}
\newcommand\HUGE{\@setfontsize\Huge{40}{50}}
\DeclareMathOperator*{\argmax}{arg\,max}

\newcommand{\alglinelabel}{%
  \addtocounter{ALC@line}{-1}
  \refstepcounter{ALC@line}
  \label
}

\DeclarePairedDelimiterX{\infdivx}[2]{(}{)}{%
  #1\;\delimsize\|\;#2%
}
\newcommand{\infdiv}{D_{\rm{KL}}\infdivx}

\def\addlegendimage{\csname pgfplots@addlegendimage\endcsname}
\definecolor{color1}{HTML}{e0ecf4}
\definecolor{color2}{HTML}{9ebcda}
\definecolor{color3}{HTML}{8856a7}

\definecolor{color4}{HTML}{f15757}
\definecolor{color5}{HTML}{6fcf97}
\definecolor{color6}{HTML}{4169e1}

\definecolor{color7}{HTML}{fde0dd}
\definecolor{color8}{HTML}{fa9fb5}
\definecolor{color9}{HTML}{c51b8a}

\definecolor{box1}{HTML}{7f7f7f}
\definecolor{box2}{HTML}{c03d3e}
\definecolor{box3}{HTML}{3b913b}
\definecolor{box4}{HTML}{3274a1}

\definecolor{azure}{HTML}{4385bc}
\definecolor{green(munsell)}{HTML}{53b251}
\definecolor{tractorred}{HTML}{e41e1e}

\usepackage{url}
\usepackage{hyperref}
\usepackage[nameinlink]{cleveref}
\creflabelformat{equation}{#2\textup{#1}#3}

\usepackage[accepted]{styles/icml2020}
\icmltitlerunning{Learning to Generate Noise for Multi-Attack Robustness}
\begin{document}
\twocolumn[
\icmltitle{Learning to Generate Noise for Multi-Attack Robustness}

\icmlsetsymbol{equal}{*}

\begin{icmlauthorlist}
\icmlauthor{Divyam Madaan}{cs}
\icmlauthor{Jinwoo Shin}{ee,ai}
\icmlauthor{Sung Ju Hwang}{cs,ai,aitrics}
\end{icmlauthorlist}

\icmlaffiliation{cs}{School of Computing, KAIST, South Korea}
\icmlaffiliation{ee}{School of Electrical Engineering, KAIST, South Korea}
\icmlaffiliation{ai}{Graduate School of AI, KAIST, South Korea}
\icmlaffiliation{aitrics}{AITRICS, South Korea}

\icmlcorrespondingauthor{Divyam~Madaan}{dmadaan@kaist.ac.kr}

\icmlkeywords{Machine Learning, ICML}

\vskip 0.3in
]

\printAffiliationsAndNotice{}
\begin{abstract}
Adversarial learning has emerged as one of the successful techniques to circumvent the susceptibility of existing methods against adversarial perturbations. However, the majority of existing defense methods are tailored to defend against a single category of adversarial perturbation (e.g. $\ell_\infty$-attack). In safety-critical applications, this makes these methods extraneous as the attacker can adopt diverse adversaries to deceive the system. Moreover, training on multiple perturbations simultaneously significantly increases the computational overhead during training. To address these challenges, we propose a novel meta-learning framework that explicitly learns to generate noise to improve the model's robustness against multiple types of attacks. Its key component is \emph{Meta Noise Generator (MNG)} that outputs optimal noise to stochastically perturb a given sample, such that it helps lower the error on diverse adversarial perturbations. By utilizing samples generated by MNG, we train a model by enforcing the label consistency across multiple perturbations. We validate the robustness of models trained by our scheme on various datasets and against a wide variety of perturbations, demonstrating that it significantly outperforms the baselines across multiple perturbations with a marginal computational cost. 
\end{abstract}

\section{Introduction}
Deep neural networks have demonstrated enormous success on multiple benchmark applications~\cite{amodei2016deep, devlin2018bert}, by achieving super-human performance on certain tasks. However, to deploy them to safety-critical applications~\cite{shen2017deep, chen15driving, mao2019metric}, we need to ensure that the model is \emph{robust} as well as \emph{accurate}, since incorrect predictions may lead to severe consequences. Notably, it is well-known that the existing neural networks are highly susceptible to \emph{adversarial examples}~\cite{szegedy2013intriguing}, which are carefully crafted image perturbations that are imperceptible to humans but derail the predictions of these otherwise accurate networks.

The emergence of adversarial examples has received significant attention in the research community, and several empirical~\cite{madry2017towards,dhillon2018stochastic,song2018pixeldefend, Zhang2019TheoreticallyPT, carmon19unlabeled, Pang2020Rethinking} and certified~\cite{wong2017provable, raghunathan2018certified,cohen2019certified} defense mechanisms have been proposed to circumvent this phenomenon. However, despite a large literature to improve upon the robustness of neural networks, most of the existing defenses leverage the knowledge of the adversaries and are based on the assumption of only a single type of adversarial perturbation. Consequently, many of the proposed defenses were circumvented by stronger attacks~\cite{carlini2017towards, anish18obfuscated,uesato2018adversarial, tramer2020adaptive}. 

{Meanwhile, several recent works~\cite{schott2018towards, tramer2019adversarial} have demonstrated the vulnerability of existing defense methods against multiple perturbations.
For the desired multi-attack robustness, \citet{tramer2019adversarial, maini2019adversarial} proposed various strategies to aggregate multiple perturbations during training. However, training with multiple perturbations comes at an additional cost; it increases the training cost by a factor of four over adversarial training, which is already an order of magnitude more costly than standard training.} This slowdown factor hinders the research progress of robustness against multiple perturbations due to the large computation overhead incurred during training. Some recent works reduce this cost by reducing the complexity of generating adversarial examples~\cite{shafahi2019free, wong2020fast}, however, they are limited to $\ell_\infty$ adversarial training. 

\begin{figure*}[t!]
    \centering
    \def\svgwidth{\linewidth}
    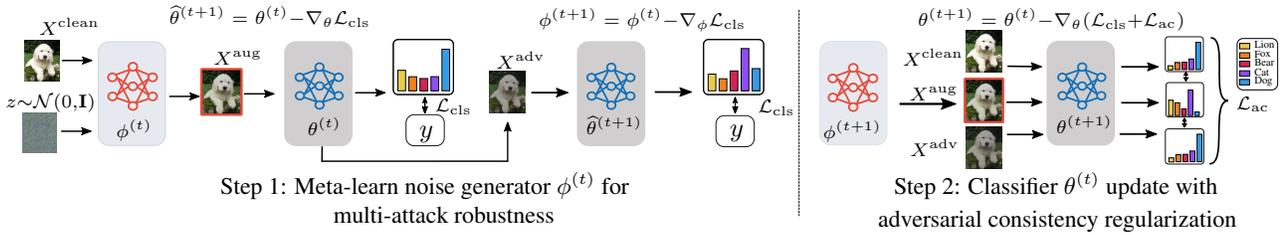

    \caption{{\bf Overview of Meta-Noise Generator with Adversarial Consistency (MNG-AC)}. The generator ${\phi}^{(t)}$ takes stochastic noise and input $X^{\rm{clean}}$ to generate the noise-augmented sample $X^{\rm{aug}}$, which is used for a temporary update of the classifier to increase the influence of the augmented examples. The generator is learnt via meta-learning by minimizing the loss on adversarial examples generated from an attack sampled uniformly from the perturbation set. The classifier $\theta^{(t)}$ then minimizes the stochastic adversarial classification loss $\calL_{\rm{cls}}$ and the adversarial consistency loss $\calL_{\rm{ac}}$. \label{fig:intro_figure}}
\end{figure*}

{To address the drawbacks of existing methods, we propose a novel training scheme, \emph{Meta Noise Generator with Adversarial Consistency (MNG-AC)}, which learns instance-dependent noise to minimize the adversarial loss across multiple perturbations while enforcing label consistency between them, as illustrated in~\Cref{fig:intro_figure} and explained in details below. 

First, we tackle the heavy computational overhead incurred by multi-perturbation training by proposing \emph{Stochastic Adversarial Training (SAT)}, that uniformly samples from
a distribution of perturbations during training, which significantly accelerates training for multiple perturbations\footnote{By a factor of four on a single machine with four GeForce RTX 2080Ti on CIFAR-10 and SVHN dataset using Wide~ResNet~28-10~\cite{zagoruyko2016wide} architecture.}.
Then, based on the assumption that the model should output the same predictions for different perturbations of the same image, we introduce \emph{Adversarial Consistency (AC)} loss that enforces label consistency across multiple perturbations. Finally,  motivated by the noise regularization techniques~\cite{huang2016deep, srivastava2014dropout, noh2017regularizing, Lee2020Meta} which target generalization, we formulate a \emph{Meta Noise Generator~(MNG)} that learns to stochastically perturb a given sample in a meta-learning framework to explicitly improve the generalization and label consistency across multiple attacks. In particular, \emph{Meta~Noise Generator with Adversarial Consistency (MNG~-~AC)} utilizes the generated samples to enforce label consistency across the generated samples from MNG, adversarial samples, and clean samples. Consequently, it increases the smoothness of the model (see~\Cref{fig:loss_plots}) and pushes the decision boundary away from the data (see~\Cref{fig:decision_boundary}) to improve the robustness across multiple perturbations.}

We extensively validate the robustness and computational efficiency of our proposed method by evaluating it on state-of-the-art attack methods and comparing it against existing state-of-the-art single and multi-perturbation adversarial defense methods on multiple benchmark datasets (CIFAR-10, SVHN, and Tiny-ImageNet dataset). The experimental results show that our method obtains significantly superior performance over all the baseline methods trained with multiple adversarial perturbations, generalizes to diverse perturbations, and substantially reduces the computational cost incurred by training with multiple adversarial perturbations.  
m
In summary, the contributions of our paper are as follows:
\begin{itemize}
\item We introduce \emph{Adversarial Consistency (AC)} loss that enforces label consistency across multiple perturbations to enforce smooth and robust networks.
\item We formulate \emph{Meta-Noise Generator (MNG)} that explicitly meta-learns an input-dependent noise generator, such that it outputs stochastic noise distribution to improve the model's robustness and adversarial consistency across multiple types of adversarial perturbations.
\item We validate our proposed method on various datasets against diverse benchmark adversarial attacks, on which it achieves state-of-the-art performance, highlighting its practical impact.
\end{itemize}

We  release  our  code  with  the pre-trained models for reproducing all the experiments at \url{https://github.com/divyam3897/MNG_AC}.

\section{Related work}\label{section:relatedWork}
{\bf Robustness against single adversarial perturbation.} In the past few years, multiple defenses have been proposed to defend against a single type of attack~\citep{madry2017towards,Xiao2020Enhancing, Zhang2019TheoreticallyPT, carmon19unlabeled, madaan2019adversarial, dongxian20weight} and have been consequently circumvented by stronger attacks~\citep{anish18obfuscated,brendel2018decisionbased, tramer2020adaptive, croce2020reliable}. Adversarial-training based defenses~\citep{madry2017towards,Zhang2019TheoreticallyPT, carmon19unlabeled, madaan2019adversarial, dongxian20weight} have been the only exceptions that have withstood the intense scrutiny and have provided empirical gains in adversarial robustness. Recently,~\citet{croce21mind} proposed $\ell_1$-APGD adversarial training to achieve better performance than MNG-AC on $\ell_1$-attack. However, we believe that the comparison is unfair, as MNG-AC does not target $\ell_1$-norm robustness only and its lower performance on $\ell_1$-norm is due to the trade-off between multiple perturbations~\citep{schott2018towards, tramer2019adversarial}. Moreover, all these prior single perturbation defenses are restricted to a single $\ell_p$-threat model and are not capable to defend against multiple perturbations simultaneously.

{\bf Robustness against multiple adversarial perturbations.} ~\citet{schott2018towards} demonstrated that $\ell_\infty$ adversarial training is highly susceptible to $\ell_0/\ell_2$-norm adversarial perturbations and used multiple VAEs to defend against multiple perturbations on the MNIST dataset. However, it was not scalable and limited to the MNIST dataset.~\citet{tramer2019adversarial} investigated the theoretical/empirical trade-offs between multiple perturbations and introduced adversarial training with worst/average perturbations to defend against multiple perturbations.~\citet{maini2019adversarial} incorporated multiple perturbations into a single adversary to maximize the adversarial loss. However, computing all the perturbations is impractical for multiple perturbations and large scale datasets. On the other hand, our proposed framework overcomes this limitation, with improved performance over these methods and has a negligible increase in training cost over multi-perturbation adversarial training.

{\bf Generative models for adversarial robustness.} There have been various attempts
that leverage the representative power of generative models to improve model robustness. \citet{samangouei2018defensegan, jalal2017robust} project an image onto the generator manifold, which is then classified by the discriminator.~\citet{song2018pixeldefend} uses the sensitivity of generative models to defend against a single perturbation. \citet{Yin2020GAT} proposed a detection method based on input space partitioning. However, \citet{samangouei2018defensegan, jalal2017robust,song2018pixeldefend} were shown to be ineffective by stronger attacks~\citep{carlini2017towards, anish18obfuscated}. In contrast to learning the generative model to model the adversarial examples, we meta-learn the generator to explicitly learn an input-dependent optimal noise distribution to lower adversarial error across multiple perturbations, that does not necessarily correspond to any of the attack perturbations.

\section{Robustness against multiple perturbations}\label{section:preliminaries}
We first briefly review single/multi-perturbation adversarial training and introduce \emph{Stochastic~Adversarial~Training (SAT)} to reduce the computational cost incurred by training with multiple perturbations. We consider a dataset $\calD$ over observations $x \in \mathbb{R}^d$ and labels $y \in \mathbb{R}^C$ with $C$ classes. Let $f_{\theta}: \mathbb{R}^d \to \mathbb{R}^C$ be a classifier with parameters $\theta$ and classification loss $\calL_{\rm{cls}}$. Given an attack procedure $\calA(x)$ {with norm-ball $\mathcal{B}(x, \varepsilon)$  around $x$ with radius $\varepsilon$ for each example,} which introduces a perturbation $\delta$, we let $x^{\rm{adv}} = x + \delta$ denote the corresponding adversarial examples. {We consider the $\ell_p$ norm attacks and adopt the projected-gradient descent (PGD)~\citep{madry2017towards} for crafting the $\ell_p$ perturbations:
\begin{equation}\label{eq:pgd_attack}
\small
x_{(t+1)}^{\rm{adv}} = \underset{\mathcal{B}(x, \varepsilon)}{\rm{proj}}\left(x_{(t)}^{\rm{adv}}+ \underset{||v||}{\argmax}~v^{T}\triangledown_{x^{\rm{adv}}_{(t)}} \mathcal{L}_{\rm{cls}}\left(f_{\theta}\left(x^{\rm{adv}}_{(t)}\right), y\right)\right),
\end{equation}
where $x^{\rm{adv}}_0$ is chosen at random within $\mathcal{B}(x, \varepsilon)$, $\alpha$ is the step size, $\rm{proj}$ is the projection operator projecting the input onto the norm ball $\mathcal{B}(x, \varepsilon)$, and $x_{(t+1)}^{\rm{adv}}$ denotes the adversarial example at the $t$-th PGD step.} We will refer the approximation of the maximum loss by an attack procedure $\calA(x)$ as $\max_{\delta \in \calB\left(x, \varepsilon\right)} \calL_{\rm{cls}}\left(f_{\theta}\left(x+ \delta\right),y\right)~\approx~\calL_{\rm{cls}}\left(f_{\theta}\left(\calA\left({x}\right)\right),y\right)$ for the rest of our paper.

{\bf Single-perturbation adversarial training.} In the standard adversarial training~\citep{kurakin2016adversarial, madry2017towards}, the model optimizes the network using a min-max formulation. More formally, the inner maximization generates the adversarial perturbation by maximizing the loss, while the outer minimization minimizes the loss on the generated examples.
\begin{equation}\label{eq:adv_madry}
\min_{\theta}~ \mathbb{E}_{(x,y) \sim \calD}~\calL_{\rm{cls}}\left(f_{\theta}\left(\calA\left({x}\right)\right), y\right).
\end{equation}

The majority of existing defenses are primarily able to defend against a single category of adversarial perturbation. However, this limits the generalization of these methods to perturbations that are unseen during training~\citep{schott2018towards, tramer2019adversarial}, which has been referred to as \emph{overfitting} on the particular type of training perturbation.

{\bf Multi-perturbation adversarial training.}~\citet{tramer2019adversarial} extended the adversarial training to multiple perturbations by optimizing the outer objective in Eq.~\eqref{eq:adv_madry} on the strongest/union of adversarial perturbations for each input example as follows:
\begin{enumerate}
\item \textbf{The maximum over all perturbations}: It optimizes the outer objective in Eq.~\eqref{eq:adv_madry} on the strongest adversarial perturbation from the perturbation set.
    \begin{equation}\label{eq:max}
        \min_{\theta}~ \mathbb{E}_{\substack{(x,y) \sim \calD}}  \biggl[{\argmax}_{k}~\calL_{\rm{cls}}\left(f_{\theta}\left(\calA_k\left(x\right)\right), y\right) \biggr].
     \end{equation}
\item \textbf{The average over all perturbations:} It optimizes the outer objective in Eq.~\eqref{eq:adv_madry} on the set of $n$ perturbations. 
  \begin{equation}\label{eq:avg}
        \min_{\theta}~ \mathbb{E}_{\substack{(x,y) \sim \calD}} \frac{1}{n}\sum_{k=1}^{k=n} \calL_{\rm{cls}}\left(f_{\theta}\left(\calA_k\left(x\right), y\right)\right).
  \end{equation}
  \end{enumerate}
Recently,~\citet{maini2019adversarial} proposed “Multi Steepest Descent” (MSD) by incorporating the different perturbations into the direction of steepest descent. However, the practicality of 
all these methods is limited due to an increased computational overhead for training.

{\bf Stochastic Adversarial Training (SAT).}  To overcome this limitation, we propose Stochastic Adversarial Training to defend against multiple adversarial perturbations. Specifically, we conjecture that it is essential to cover the threat model during training, not utilizing all the perturbations simultaneously. We formulate the threat model as a random attack $\calA(x)$ sampled uniformly from a perturbation set $S$ during each episode (or batch) of training which prevents overfitting on a particular adversarial perturbation. In this work, we consider the $\ell_p$-bounded perturbation set, and we sample the attack procedure $\calA(x)$ from the perturbation set $S$ as follows:
\begin{align}\label{eq:attack_sample}
    S &= {\{\calA_1(x),\ldots,\calA_n(x)\}}, \nonumber \\
    k &\sim {\rm{Cat}}\left(\left(1/n,\ldots, 1/n\right)\right), \nonumber \\
    \calA(x) &= S_k(x),
\end{align}
where $\rm{Cat}$ is the categorical distribution and $n$ is the number of attacks in the perturbation set $S$. Our proposed SAT optimizes the outer objective in Eq.~\eqref{eq:adv_madry} using the sampled attack procedure $\calA(x)$ and is a drastic simplification of the average strategy in Eq.~\eqref{eq:avg}, which makes it highly efficient for multiple perturbations. It is important to note that unlike the average and max strategy SAT can be applied to any perturbation set with a constant cost and it promotes generalization and convergence (due to its stochasticity) by preventing over-fitting on a single type of perturbation.
\section{Learning to generate noise for multi-attack robustness}\label{section:proposedMethod}
In this section, we introduce our framework MNG-AC, which leverages an \emph{adversarial consistency loss (AC)} and a \emph{meta-noise generator (MNG)} to help the model generalize to multiple perturbations. Let $x^{\rm{adv}}_{\theta}$ be the adversarial examples generated from the network $f_{\theta}$ for a uniformly sampled attack $\calA(x)$ with norm-ball $\mathcal{B}(x, \varepsilon)$ from a perturbation set $S$ and $g_{\phi}: \mathbb{R}^{d} \to \mathbb{R}^d$ denotes the generator with parameters $\phi$. We input $z\sim\calN(0,\bf{I})$ to our generator jointly with the clean examples $x$ to generate the noise-augmented samples $x^{\rm{aug}}_{\phi}$ projected on the same norm-ball $\calB(x, \varepsilon)$ as:
\begin{equation}\label{eq:aug_examples}
        x^{\rm{aug}}_{\phi} = \underset{\calB(x, \varepsilon)}{\rm{proj}}\left(x + g_{\phi}(z, x)\right),~~\text{where}~~ z\sim \calN(0, \bf{I}).
\end{equation}
The total loss function $\calL_{\rm{total}}$ for the classifier consists exclusively of two terms: SAT classification loss and an adversarial consistency loss:
\begin{equation}\label{eq:total_loss}
\begin{aligned}
   \calL_{\rm{total}} &= \frac{1}{B}\sum_{i=1}^{B}~~\underbrace{\mathcal{L}_{\rm{cls}}\left(\theta\mid {x^{\rm{adv}}_\theta(i), y(i)}\right)}_{\text{SAT classification loss}} \\ &+ \underbrace{\beta \cdot {\calL_{\rm{ac}}}\left({p^{\rm{clean}}(i); p^{\rm{adv}}(i); p^{\rm{aug}}(i)}\right)}_{\text{adversarial consistency loss}}, \\
\end{aligned}
\end{equation}
where $B$ is the batch-size, $\beta$ is the hyper-parameter determining the strength of the AC loss denoted by $\calL_{\rm{ac}}$ and $p^{\rm{clean}}, {p}^{\rm{adv}}, {p}^{\rm{aug}}$ represent the posterior distributions ${p}(y\mid x^{\rm{clean}}), {p}(y\mid {x^{\rm{adv}}_{\theta}}), {p}(y\mid {x^{\rm{aug}}_{\phi}})$ {computed using the softmax function on the logits for $x^{\rm{clean}}$, $x^{\rm{adv}}$, and $x^{\rm{aug}}$ respectively. Specifically, $\calL_{\rm{ac}}$ represents the Jensen-Shannon Divergence (JSD) among the posterior distributions:
\begin{align}
   \calL_{\rm{ac}} &= \frac{1}{3}\left(\infdiv{p^{\rm{clean}}}{M} + \infdiv{p^{\rm{adv}}}{M}\right. \nonumber \\ &+ \left.\infdiv{p^{\rm{aug}}}{M} \right),
\end{align}
where $M = \left(p^{\rm{clean}} + p^{\rm{adv}} + p^{\rm{aug}}\right)/3$.} Consequently, $\calL_{\rm{ac}}$ enforces stability and insensitivity across a diverse range of inputs based on the assumption that the classifier should output similar predictions when fed perturbed versions of the same image.

\begin{algorithm}[t]
\caption{Algorithm for MNG-AC}
\label{alg:method}
\begin{algorithmic}[1]
\INPUT Dataset $\calD$, $T$ epochs, batch size $B$, perturbation set~$S$, classifier $\theta$ and noise-generator $\phi$.
\OUTPUT  Final model parameters $\theta$.
\FOR{\text{$t = \{1, \dots, T\}$} \label{alg:step1}}
\STATE{\text{Sample mini-batch of size $B$.}} \label{alg:step2}
\STATE{\text{{Sample an attack procedure $\calA(x)$ from $S$ (Eq.~\eqref{eq:attack_sample}).}}}\alglinelabel{alg:step3}
\STATE{\text{{Generate adversarial examples for $\calA(x)$ (Eq.~\eqref{eq:pgd_attack}}}}).
\STATE{\text{Generate ${x^{\rm{aug}}_{\phi}}$ using $\phi^{(t)}$ by Eq.~\eqref{eq:aug_examples}}}.
\STATE{\text{Temporary update of $\theta^{(t)}$ using Eq.~\eqref{eq:first_step}}}.
\STATE{\text{Update the parameters $\phi^{(t)}$ of MNG by Eq.~\eqref{eq:second_step}}}.
\STATE{\text{Generate ${x^{\rm{aug}}_{\phi}}$ using $\phi^{(t+1)}$ by Eq.~\eqref{eq:aug_examples}}}.
\STATE{\text{Update $\theta^{(t)}$ by Eq. \eqref{eq:third_step}}}.

\ENDFOR
\end{algorithmic}
\end{algorithm}

Recently,~\citet{rusak2020increasing} formulated an adversarial noise generator to learn the adversarial noise to improve the robustness on common corruptions. However, our goal is different; the robustness against multiple adversarial attacks is a much more challenging task than that against common corruptions.
To generate the augmented samples for our purpose, MNG meta-learns~\citep{thrun98, finn2017model} the parameters $\phi$ of the noise generator $g_{\phi}$ to generate an input-dependent noise distribution to alleviate the issue of generalization across multiple adversaries. The standard approach to train our adversarial classifier jointly with MNG is to use bi-level optimization~\citep{finn2017model}. However, bi-level optimization for adversarial training would be computationally expensive. 

To tackle this challenge, we adopt an online approximation~\citep{ren2018learning, shu2019meta} to update $\theta$ and $\phi$ using a single-optimization loop. We alternatively update the parameters $\theta$ of the classifier with the parameters $\phi$ of MNG using the following training scheme: 

\begin{enumerate}
\item \textbf{Temporary model update on augmented samples.} First, we update $\theta$ to minimize $\calL_{\rm{cls}}(\theta \mid {x^{\rm{aug}}_\phi}, y)$, which ensures the learning of the classifier using the generated samples constructed by MNG. It explicitly increases the influence of the noise-augmented samples on the classifier. More specifically, for a learning rate $\alpha$, projection operator $\rm{proj}$, current $\theta^{(t)}$ moves along the following descent direction:
\begin{equation}\label{eq:first_step}
    \begin{aligned}
        \widehat{\theta}^{(t)} = \theta^{(t)} - &\alpha\cdot\frac{1}{B}\sum_{i=1}^{B} \nabla_\theta \calL_{\rm{cls}}\left(\theta^{(t)}\mid {x^{\rm{aug}}_\phi(i), y(i)}\right). \\
    \end{aligned}
\end{equation}  

\item \textbf{Update generator parameters.} After receiving feedback from the classifier, we adapt $\phi$ to minimize the SAT loss (adversarial loss on the uniformly sampled attack $\calA(x)$). In particular, $\phi$ facilitates the classifier parameters $\theta$ in the next step with the update step\footnote{Note that $\phi$ is a variable in this case, which makes the loss in Eq.~\ref{eq:second_step} a function of $\phi$, allowing the the gradients' computation.}:
\begin{equation}\label{eq:second_step}
 \phi^{(t+1)} = \phi^{(t)} - \alpha\cdot\frac{1}{B}\sum_{i=1}^{B}\nabla_{\phi}\calL_{\rm{cls}}\left(\widehat{\theta}^{(t)}\mid {x^{\rm{adv}}_\theta(i), y(i)}\right).
\end{equation}

\item \textbf{Update model parameters.} Finally, we update $\theta^{(t)}$ to minimize loss from Eq.~\eqref{eq:total_loss}. This step explicitly models the adaptation of adversarial model parameters in the presence of the noise-augmented data using the adversarial consistency loss:
    \begin{align}\label{eq:third_step}
    \theta^{(t+1)} = \theta^{(t)} &- \frac{1}{B} \sum^{B}_{i=1} \left(\mathcal{L}_{\rm{cls}}\left(\theta\mid {x^{\rm{adv}}_\theta(i), y(i)}\right)\right. \nonumber \\ &+ \left.\beta \cdot {\calL_{\rm{ac}}}\left({p^{\rm{clean}}(i); p^{\rm{adv}}(i); p^{\rm{aug}}(i)}\right)\right).
        \end{align}
\end{enumerate}  

To summarize, MNG-AC utilizes perturbation sampling to generate the adversarial examples. The generator perturbs the clean examples in a meta-learning framework to explicitly lower the loss on the generated adversarial examples. Lastly, the adversarial classifier utilizes the generated samples, adversarial samples and clean samples to optimize the adversarial classification and adversarial consistency loss. 

{\bf Intuition behind our framework.}
Unlike existing defenses that aim for robustness against a single perturbation, our proposed approach targets for a realistic scenario of robustness against multiple perturbations. Our motivation is that meta-learning the noise distribution to minimize the SAT loss allows to learn the optimal noise to improve multi-perturbation generalization. Based on the assumption that the model should output similar predictions for perturbed versions of the same image, we enforce the AC loss, which enforces the label consistency across multiple perturbations. 
\section{Experiments}\label{section:experiments} 
\begin{table*}[ht!]
\centering
\small
\renewcommand{\arraystretch}{1.1}
	\begin{minipage}[b]{\linewidth}
		\captionof{table}{Comparison  of robustness against multiple types of perturbations. 
		All the values are measured by computing mean, and standard deviation across three trials, the best and second-best results are highlighted in {\bf bold} and \underline{underline} respectively. Time denotes the training time in hours. {For CIFAR-10 and SVHN, we use $\varepsilon = \{\frac{8}{255}, \frac{2000}{255}, \frac{128}{255}\}$ for $\ell_\infty, \ell_1$, and $\ell_2$ attacks respectively. For Tiny-ImageNet, we use $\varepsilon = \{\frac{4}{255}, \frac{2000}{255}, \frac{80}{255}\}$ for $\ell_\infty, \ell_1$, and $\ell_2$ attacks respectively.} We report the worst-case accuracy for all the attacks and defer the breakdown of all attacks to \Cref{section:appendix_experiments}.
	\label{table:results}}
\resizebox{\linewidth}{!}{
\begin{tabular}{@{}llccccccc@{}}
\toprule
& Model & Acc$_{\rm{clean}}$ & $\ell_\infty$ & $\ell_1$ & $\ell_2$  & Acc$^{\rm{union}}_{\rm{adv}}$ & Acc$^{\rm{avg}}_{\rm{adv}}$ & Time (h) \\ \midrule
\parbox[t]{2mm}{\multirow{10}{*}{\rotatebox[origin=c]{90}{CIFAR-10}}}
& Nat~{\scriptsize{\citep{zagoruyko2016wide}}} & {\bf 94.7{\scriptsize $\pm$ 0.1}} & 0.0{\scriptsize $\pm$ 0.0} & 0.0{\scriptsize $\pm$ 0.0} & 0.4{\scriptsize $\pm$ 0.2} &  0.0{\scriptsize $\pm$ 0.0} & 0.0{\scriptsize $\pm$ 0.0} & {\bf 0.4} \\
& Adv$_\infty$~\scriptsize{\citep{madry2017towards}} &  86.8{\scriptsize $\pm$ 0.1} & \underline{44.9{\scriptsize $\pm$ 0.7}}  & 26.2{\scriptsize $\pm$ 0.4}  & 55.0{\scriptsize $\pm$ 0.9} & 25.6{\scriptsize $\pm$ 0.6} & 41.9{\scriptsize $\pm$ 0.6} & 4.5\\
& Adv$_1$ & {93.3{\scriptsize $\pm$ 0.4}} & 0.0{\scriptsize $\pm$ 0.0}  & {\bf 80.7{\scriptsize $\pm$ 0.7}} & {0.0\scriptsize $\pm$ 0.0} & 0.0{\scriptsize $\pm$ 0.00} &  26.8{\scriptsize $\pm$ 0.6} & 8.1\\
& Adv$_2$ & {89.4{\scriptsize $\pm$ 0.2}}  & 28.8{\scriptsize $\pm$ 1.3}  & 54.2{\scriptsize $\pm$ 0.4}  & {\bf 65.8{\scriptsize $\pm$ 0.3}} &  28.6{\scriptsize $\pm$ 1.4} & 49.6{\scriptsize $\pm$ 0.3} & 3.7 \\
& TRADES$_\infty$~\scriptsize{\citep{Zhang2019TheoreticallyPT}} & 84.7{\scriptsize $\pm$ 0.3}  & {\bf 48.9{\scriptsize $\pm$ 0.7}}  & 32.3{\scriptsize $\pm$ 1.0}  & 57.8{\scriptsize $\pm$ 0.6} & 31.5{\scriptsize $\pm$ 1.2} & 46.3{\scriptsize $\pm$ 0.7} & 5.2 \\
\cmidrule{2-9}
& Adv$_{\text{avg}}$~\scriptsize{\citep{tramer2019adversarial}} &  86.0{\scriptsize $\pm$ 0.1}  & 34.1{\scriptsize $\pm$ 0.5}  & 61.3{\scriptsize $\pm$ 0.6}  & \underline{65.7{\scriptsize $\pm$ 0.4}} & 34.1{\scriptsize $\pm$ 0.1} & {53.7{\scriptsize $\pm$ 0.3}} & 16.9\\
& Adv$_{\text{max}}$~\scriptsize{\citep{tramer2019adversarial}} & 84.2{\scriptsize $\pm$ 0.1} & 39.9{\scriptsize $\pm$ 0.5}  & { 57.9{\scriptsize $\pm$ 0.7}} & 64.5{\scriptsize $\pm$ 0.1} & {39.7{\scriptsize $\pm$ 0.5}} & \underline{54.1{\scriptsize $\pm$ 0.4}} & 16.3\\
& MSD~\scriptsize{\citep{maini2019adversarial}} & 82.7{\scriptsize $\pm$ 0.1} & 43.5{\scriptsize $\pm$ 0.5} & 54.3{\scriptsize $\pm$ 0.4} & 63.1{\scriptsize $\pm$ 0.5} & {\bf 42.7{\scriptsize $\pm$ 0.5}} & 53.6{\scriptsize $\pm$ 0.2} & 16.7 \\
& ANT~\scriptsize{\citep{rusak2020increasing}} & \underline{94.6{\scriptsize $\pm$ 0.0}} & 0.0{\scriptsize $\pm$ 0.0} & 0.0{\scriptsize $\pm$ 0.0} & 0.0{\scriptsize $\pm$ 0.0} & 0.0{\scriptsize $\pm$ 0.0} & 0.0{\scriptsize $\pm$ 0.0} & \underline{0.7}\\
\cmidrule{2-9}
& MNG-AC (Ours) & 81.7{\scriptsize $\pm$ 0.3}  & 41.4{\scriptsize $\pm$ 0.7}  & \underline{65.4{\scriptsize $\pm$ 0.3}}  & {65.2\scriptsize $\pm$ 0.5} & \underline{41.4{\scriptsize $\pm$ 0.7}} & {\bf 57.2{\scriptsize $\pm$ 0.4}} & 8.4\\ 
\midrule

\parbox[t]{2mm}{\multirow{10}{*}{\rotatebox[origin=c]{90}{SVHN}}}
& Nat~{\scriptsize{\citep{zagoruyko2016wide}}} & {\bf 96.8{\scriptsize $\pm$ 0.1}} &  0.0{\scriptsize $\pm$ 0.0} &  9.4{\scriptsize $\pm$ 0.5} & 3.8{\scriptsize $\pm$ 0.7}  & 0.0{\scriptsize $\pm$ 0.0} & 4.5{\scriptsize $\pm$ 0.2} & {\bf 0.6} \\
& Adv$_\infty$~\scriptsize{\citep{madry2017towards}} & 92.8{\scriptsize $\pm$ 0.2} & \underline{46.2{\scriptsize $\pm$ 0.6}} & 8.2{\scriptsize $\pm$ 0.9} & 30.2{\scriptsize $\pm$ 0.5} & 8.1{\scriptsize $\pm$ 0.9} & 28.3{\scriptsize $\pm$ 0.1} &  {6.2} \\
& Adv$_1$  & 92.4{\scriptsize $\pm$ 0.9} & 0.0{\scriptsize $\pm$ 0.0} & {\bf 77.2{\scriptsize $\pm$ 2.9}} & 0.0{\scriptsize $\pm$ 0.0} & 0.0{\scriptsize $\pm$ 0.0} & 25.7{\scriptsize $\pm$ 1.0} &11.8 \\
& Adv$_2$ & {93.0{\scriptsize $\pm$ 0.1}} & 21.7{\scriptsize $\pm$ 0.4} & 44.7{\scriptsize $\pm$ 0.5} & \underline{62.9{\scriptsize $\pm$ 0.2}} & 21.0{\scriptsize $\pm$ 0.4} & 43.1{\scriptsize $\pm$ 0.3} & { 6.1}\\
& TRADES$_\infty$~\scriptsize{\citep{Zhang2019TheoreticallyPT}} & 93.9{\scriptsize $\pm$ 0.1} & {\bf 49.9{\scriptsize $\pm$ 1.7}} & 4.2{\scriptsize $\pm$ 0.4} & 26.7{\scriptsize $\pm$ 2.0} & 4.1{\scriptsize $\pm$ 0.4} & 26.9{\scriptsize $\pm$ 1.1} & 7.9 \\
\cmidrule{2-9}
& Adv$_{\text{avg}}$\scriptsize{\citep{tramer2019adversarial}} & 91.6{\scriptsize $\pm$ 0.3} & 21.5{\scriptsize $\pm$ 2.7} & {61.2{\scriptsize $\pm$ 4.1}} & 56.1{\scriptsize $\pm$ 2.3} & {20.4{\scriptsize $\pm$ 2.7}} & \underline{45.9{\scriptsize $\pm$ 0.9}} & 24.1\\
& Adv$_{\text{max}}$~\scriptsize{\citep{tramer2019adversarial}} & 86.9{\scriptsize $\pm$ 0.3} & 28.8{\scriptsize $\pm$ 0.2} & 48.9{\scriptsize $\pm$ 0.9} & 56.3{\scriptsize $\pm$ 0.8} & 28.8{\scriptsize $\pm$ 0.2} & 44.7{\scriptsize $\pm$ 0.4} & 22.7\\ 
& MSD~\scriptsize{\citep{maini2019adversarial}} & 81.8{\scriptsize $\pm$ 0.3} & 34.1{\scriptsize $\pm$ 0.3} & 43.4{\scriptsize $\pm$ 0.5} & 54.1{\scriptsize $\pm$ 0.2} & \underline{34.1{\scriptsize $\pm$ 0.3}} & 44.0{\scriptsize $\pm$ 0.1} & 23.7 \\
& ANT~\scriptsize{\citep{rusak2020increasing}}   & \underline{ 96.7{\scriptsize $\pm$ 0.0}} & 0.2{\scriptsize $\pm$ 0.0} & 15.6{\scriptsize $\pm$ 0.6} & 7.7{\scriptsize $\pm$ 0.6} & 0.0{\scriptsize $\pm$ 0.0} & 7.8{\scriptsize $\pm$ 0.4} & \underline{1.1}\\
\cmidrule{2-9}
& MNG-AC (Ours) & 92.6{\scriptsize $\pm$ 0.2} & {34.2{\scriptsize $\pm$ 1.0}} & \underline{ 71.3{\scriptsize $\pm$ 1.7}} & {\bf 66.7 \scriptsize $\pm$ 0.9} & {\bf 34.2{\scriptsize $\pm$ 1.0}} & \underline{\bf 57.4{\scriptsize $\pm$ 0.4}} & 11.9 \\
\midrule
\parbox[t]{2mm}{\multirow{10}{*}{\rotatebox[origin=c]{90}{Tiny-ImageNet}}}
& Nat~\scriptsize{\citep{he2016identity}} & {\bf 62.8\scriptsize $\pm$ 0.4} & {0.0\scriptsize $\pm$ 0.0} & {2.7\scriptsize $\pm$ 0.3} & {12.6\scriptsize $\pm$ 0.8} & {0.0\scriptsize $\pm$ 0.0} & {5.1\scriptsize $\pm$ 0.4} & {\bf 0.9}\\
& Adv$_\infty$~\scriptsize{\citep{madry2017towards}} &  54.2{\scriptsize $\pm$ 0.4} & \underline{29.6{\scriptsize $\pm$ 0.1}} & {38.2\scriptsize $\pm$ 0.7} & 42.5{\scriptsize $\pm$ 0.6}& 29.4{\scriptsize $\pm$ 0.1} & 36.7{\scriptsize $\pm$ 0.4} & 4.3\\
& Adv$_1$  & 57.8{\scriptsize $\pm$ 0.2} & 10.5{\scriptsize $\pm$ 0.7} & \underline{44.6{\scriptsize $\pm$ 0.1}} &  41.9{\scriptsize $\pm$ 0.0} &  10.1{\scriptsize $\pm$ 0.7} &  32.2{\scriptsize $\pm$ 0.4} & 12.9\\
& Adv$_2$ & {59.5{\scriptsize $\pm$ 0.1}} & 5.2{\scriptsize $\pm$ 0.6} & 44.1{\scriptsize $\pm$ 0.4} & {\bf 44.9{\scriptsize $\pm$ 0.1}} &5.2{\scriptsize $\pm$ 0.6} & 31.7{\scriptsize $\pm$ 0.5} & {3.7}\\
& TRADES$_\infty$~\scriptsize{\citep{Zhang2019TheoreticallyPT}} &  48.2{\scriptsize $\pm$ 0.2} & 28.7{\scriptsize $\pm$ 0.9} &  33.2{\scriptsize $\pm$ 0.4} &  35.8{\scriptsize $\pm$ 0.7}  &  26.1{\scriptsize $\pm$ 0.9} &  32.8{\scriptsize $\pm$ 0.1} & 5.8\\
\cmidrule{2-9}
& Adv$_{\text{avg}}$~\scriptsize{\citep{tramer2019adversarial}} & 56.0{\scriptsize $\pm$ 0.0} & 23.7{\scriptsize $\pm$ 0.2} & {43.3{\scriptsize $\pm$ 0.5}} & \underline{46.6{\scriptsize $\pm$ 1.8}} & 23.6{\scriptsize $\pm$ 0.3} & {37.2{\scriptsize $\pm$ 0.2}} & 26.8\\
& Adv$_{\text{max}}$~\scriptsize{\citep{tramer2019adversarial}} & 53.5{\scriptsize $\pm$ 0.0} & {\bf 29.8{\scriptsize $\pm$ 0.1}} & 39.5{\scriptsize $\pm$ 0.4} & 42.4{\scriptsize $\pm$ 1.0}& {\bf 29.8{\scriptsize $\pm$ 0.3}} & \underline{37.3{\scriptsize $\pm$ 0.4}} & 24.4\\ 
& MSD~\scriptsize{\citep{maini2019adversarial}} & 53.0{\scriptsize $\pm$ 0.1}& 29.4{\scriptsize $\pm$ 0.3} & {39.2{\scriptsize $\pm$ 0.1}} & 37.2{\scriptsize $\pm$ 0.1} &  \underline{29.4{\scriptsize $\pm$ 0.3}} & 35.3{\scriptsize $\pm$ 0.2} & 25.2\\
& ANT~\scriptsize{\citep{rusak2020increasing}} & \underline{62.8{\scriptsize $\pm$ 0.0}} &  0.2{\scriptsize $\pm$ 0.0} &  3.4{\scriptsize $\pm$ 0.1} &  13.4{\scriptsize $\pm$ 0.3} &  0.0{\scriptsize $\pm$ 0.0} &  5.6{\scriptsize $\pm$ 0.1} & \underline{1.2}\\
\cmidrule{2-9}
& MNG-AC (Ours) & 53.1{\scriptsize $\pm$ 0.3} &  28.1{\scriptsize $\pm$ 0.7} &  {\bf 45.1{\scriptsize $\pm$ 0.5}} & {44.4{\scriptsize $\pm$ 0.1}} & 28.1{\scriptsize $\pm$ 0.8} &  {\bf 39.1{\scriptsize $\pm$ 0.6}} & 13.2\\ 
\bottomrule
\end{tabular}}
	\end{minipage}
\end{table*}
\subsection{Experimental setup}

{\bf Datasets.}
We evaluate on multiple benchmark datasets:
\begin{enumerate}
\item \textbf{CIFAR-10.} This dataset~\citep{alex12cifar} contains 60,000 images with 5,000 images for training and 1,000 images for test for each class. Each image is sized \(32\times32\), we use the Wide ResNet 28-10 architecture~\citep{zagoruyko2016wide} as a base network for this dataset.
\item \textbf{SVHN.} This dataset~\citep{netzer2011reading} contains 73257 training and 26032 testing images of digits and numbers in natural scene images containing ten-digit classes. Each image is sized \(32\times32\), and we use the Wide ResNet 28-10 architecture similar to the CIFAR-10 dataset as the base network for this dataset.
\item \textbf{Tiny-ImageNet.} This dataset\footnote{\url{https://tiny-imagenet.herokuapp.com/}} is a subset of ImageNet~\citep{russakovsky2015imagenet} dataset, consisting of $500$, $50$, and $50$ images for training, validation, and test dataset, respectively. This dataset contains  $64 \times 64$ size images from $200$ classes, we use ResNet50~\citep{he2016identity} as a base network for this dataset.
\end{enumerate}

{\bf Baselines and our model.} 
We compare MNG-AC with the standard network (Nat) and single-perturbation baselines including \citet{madry2017towards}~(Adv$_{\rm{p}}$) for $\ell_\infty, \ell_1$, and $\ell_2$ norm, TRADES$_{\infty}$~\cite{Zhang2019TheoreticallyPT}  for $\ell_\infty$ norm. We consider state-of-the-art multi-perturbation baselines: namely, we consider Adversarial training with the maximum (see Eq.~\eqref{eq:max}) (Adv$_{\rm{max}}$), average (Adv$_{\rm{avg}}$) ~\citep{tramer2019adversarial} (see Eq.~\eqref{eq:avg}) strategies, and Multiple steepest descent (MSD)~\citep{maini2019adversarial}. 
We additionally consider Adversarial Noise Training~\citep{rusak2020increasing} that learns adversarial noise to improve robustness against common corruptions~\cite{hendrycks2018benchmarking}.

{\bf Evaluation setup.}
We have evaluated the proposed defense scheme and baselines
against perturbations generated by state-of-the-art attack methods. We validate the clean accuracy (Acc$_{\rm{clean}}$), 
the worst (Acc$^{\rm{union}}_{\rm{adv}}$) and average (Acc$^{\rm{avg}}_{\rm{adv}}$) adversarial accuracy across all the perturbation sets for all the models. For $\ell_\infty$ attacks, we use PGD~\citep{madry2017towards}, Brendel and Bethge~\citep{brendel2019nips}, and AutoAttack~\citep{croce2020reliable}. For $\ell_2$ attacks, we use CarliniWagner~\citep{carlini2017towards}, PGD~\citep{madry2017towards}, Brendel and Bethge~\citep{brendel2019nips}, and AutoAttack~\citep{croce2020reliable}. For $\ell_1$ attacks, we use SLIDE~\citep{tramer2019adversarial}, Salt and pepper~\citep{rauber2017foolbox}, and EAD attack~\citep{chen2018ead}. We provide a detailed description of the training and evaluation setup in \Cref{section:appendix_setup}. 

\subsection{Comparison of robustness against multiple perturbations}
{\bf Results with CIFAR-10 dataset.}
\Cref{table:results} shows the experimental results for the CIFAR-10 dataset. It is evident from the results that MNG-AC achieves a relative improvement of $\sim 31\%$ and $\sim  24\%$ on the Acc$^{\rm{union}}_{\rm{adv}}$ and Acc$^{\rm{avg}}_{\rm{adv}}$ metric over the best single-perturbation adversarial training methods. Furthermore, MNG-AC achieves a relative improvement of $\sim  6\%$ on the Acc$^{\rm{avg}}_{\rm{adv}}$ metric over the state-of-the-art methods trained on multiple perturbations. Moreover, MNG-AC achieves $\sim50\%$ reduction in training time compared to all the multi-perturbations training baselines. 
It is also worth mentioning that, MNG-AC also shows an improvement over Adv$_{\rm{max}}$, which is fundamentally designed to defend against the worst perturbation in the perturbation set. 

\begin{table*}[ht!]
\footnotesize
\centering
\renewcommand{\arraystretch}{1.1}
\caption{Robustness evaluation using semi-supervised learning against multiple perturbations.
All the values are measured by computing mean, and standard deviation across three trials, the best results are highlighted in {\bf bold}. Due to the computational constraints, we use efficient training techniques~\cite{smith2017cyclical, wong2020fast} for training all the methods, which result in a slightly lower performance compared to the results in the original paper~\cite{carmon19unlabeled}.\label{table:rst}}
\resizebox{\linewidth}{!}{
\begin{tabular}{@{}llccccccc@{}}
\toprule
& Model & Acc$_{\rm{clean}}$ & $\ell_\infty$ & $\ell_1$ & $\ell_2$  & Acc$^{\rm{union}}_{\rm{adv}}$ & Acc$^{\rm{avg}}_{\rm{adv}}$ & Time (h) \\ \midrule
\parbox[t]{0.4mm}{\multirow{3}{*}{\rotatebox[origin=c]{90}{{\bf \tiny CIFAR-10}}}}
& RST$_\infty$~\scriptsize{\citep{carmon19unlabeled}} & {\bf 88.9{\scriptsize $\pm$ 0.2}}  & {\bf 54.9{\scriptsize $\pm$ 1.8}}  & 36.0{\scriptsize $\pm$ 0.9}  & {59.5{\scriptsize $\pm$ 0.2}} & 35.7{\scriptsize $\pm$ 0.6} & 50.1{\scriptsize $\pm$ 0.8} & 73.5  \\
& MNG-AC (Ours) & 81.7{\scriptsize $\pm$ 0.3}  & 41.4{\scriptsize $\pm$ 0.7}  & {65.4{\scriptsize $\pm$ 0.3}}  & {65.2\scriptsize $\pm$ 0.5} & {41.4{\scriptsize $\pm$ 0.7}} & {57.2{\scriptsize $\pm$ 0.4}} & 8.4\\ 
& MNG-AC + RST (Ours) & {88.7{\scriptsize $\pm$ 0.2}} & 47.2{\scriptsize $\pm$ 0.8} & {\bf 73.8{\scriptsize $\pm$ 0.7}} & \bf{73.7{\scriptsize $\pm$ 0.2}} &  {\bf 47.2{\scriptsize $\pm$ 0.7}} & {\bf 64.9{\scriptsize $\pm$ 0.3}} & 78.5\\
\midrule

\parbox[t]{0.4mm}{\multirow{3}{*}{\rotatebox[origin=c]{90}{{\bf \tiny SVHN}}}}
& RST$_\infty$~\scriptsize{\citep{carmon19unlabeled}} & {95.6{\scriptsize $\pm$ 0.0}} & {\bf 60.9{\scriptsize $\pm$ 2.0}} & 3.5{\scriptsize $\pm$ 0.5} & 28.8{\scriptsize $\pm$ 0.9} & 3.5{\scriptsize $\pm$ 0.5} & 31.1{\scriptsize $\pm$ 0.6} & 81.0 \\
& MNG-AC (Ours) & 92.6{\scriptsize $\pm$ 0.2} & 34.2{\scriptsize $\pm$ 1.0} & 71.3{\scriptsize $\pm$ 1.7} & {{66.7 \scriptsize $\pm$ 0.9}} & {34.2{\scriptsize $\pm$ 1.0}} & {57.4{\scriptsize $\pm$ 0.4}} & 11.9 \\
& MNG-AC + RST (Ours) & {\bf 96.3{\scriptsize $\pm$ 0.3}} & 43.8{\scriptsize $\pm$ 1.5} & {\bf 78.9{\scriptsize $\pm$ 2.0}} & {\bf 72.6{\scriptsize $\pm$ 0.2}} & {\bf 43.8{\scriptsize $\pm$ 1.5}} & {\bf 65.1{\scriptsize $\pm$ 0.4}} & 85.0\\
\bottomrule
\end{tabular}}
\vspace{-0.1in}
\end{table*}
\begin{table*}[t!]
\footnotesize
\centering
\renewcommand{\arraystretch}{1.1}
\captionof{table}{Ablation study analyzing the significance of SAT, Adversarial Consistency loss (AC) and Meta Noise Generator (MNG). The best results are highlighted in {\bf bold}.\label{tab:ablation_table}} 
\resizebox{\linewidth}{!}{
\begin{tabular}{@{}rccc|lcccccccc@{}}
\toprule
& SAT & AC & MNG & Acc$_{\rm{clean}}$ & $\ell_\infty$ & $\ell_1$ & $\ell_2$  &  Acc$^{\rm{union}}_{\rm{adv}}$ & Acc$^{\rm{avg}}_{\rm{adv}}$ & Time (h) \\ \midrule
\parbox[t]{0.35mm}{\multirow{3}{*}{\rotatebox[origin=c]{90}{{\bf \tiny CIFAR-10}}}}
& $\checkmark$ & - & - & {\bf 86.6{\scriptsize $\pm$ 0.0}} & 35.1{\scriptsize $\pm$ 0.5} & 61.8{\scriptsize $\pm$ 1.1} & {\bf 66.9{\scriptsize $\pm$ 0.4}} & 35.0{\scriptsize $\pm$ 0.5} & {54.6{\scriptsize $\pm$ 0.2}} & {\bf 5.5}\\
&$\checkmark$ & $\checkmark$ & - & 80.3{\scriptsize $\pm$ 0.2}  & 40.6{\scriptsize $\pm$ 0.8} & 62.0{\scriptsize $\pm$ 0.2} & 63.5{\scriptsize $\pm$ 0.6} & 40.6{\scriptsize $\pm$ 0.1} & 55.4{\scriptsize $\pm$ 0.1} & 6.8\\
&$\checkmark$ & $\checkmark$ & $\checkmark$ & 81.7{\scriptsize $\pm$ 0.3}  & {\bf 41.4{\scriptsize $\pm$ 0.7}}  & {\bf 65.2{\scriptsize $\pm$ 0.3}}  & {65.4\scriptsize $\pm$ 0.5} & {\bf 41.4{\scriptsize $\pm$ 0.7}} & {\bf 57.2{\scriptsize $\pm$ 0.4}} & 8.4 \\ 
\midrule
\parbox[t]{0.4mm}{\multirow{3}{*}{\rotatebox[origin=c]{90}{\bf \tiny SVHN}}}
&$\checkmark$ & - & - & 92.3{\scriptsize $\pm$ 0.1} & 26.2{\scriptsize $\pm$ 0.8} & {64.4{\scriptsize $\pm$ 0.2}} & 63.2{\scriptsize $\pm$ 0.8} & 26.2{\scriptsize $\pm$ 0.8} & {51.0{\scriptsize $\pm$ 0.1}} & {\bf 7.6} \\
&$\checkmark$ & $\checkmark$ & - & 92.2{\scriptsize $\pm$ 0.3} & 31.4{\scriptsize $\pm$ 1.3} & 65.2{\scriptsize $\pm$ 3.6} & 63.9{\scriptsize $\pm$ 0.5} & 31.1{\scriptsize $\pm$ 1.4} & 53.5{\scriptsize $\pm$ 0.8} & 8.7\\ 
&$\checkmark$ & $\checkmark$ & $\checkmark$ & {\bf 92.6{\scriptsize $\pm$ 0.2}} & {\bf 34.2{\scriptsize $\pm$ 1.0}} & {\bf 71.3{\scriptsize $\pm$ 1.7}} & {\bf {66.7 \scriptsize $\pm$ 0.9}} & {\bf 34.2{\scriptsize $\pm$ 1.0}} & {\bf 57.4{\scriptsize $\pm$ 0.4}} & 11.9 \\ 
\bottomrule
\end{tabular}}
\end{table*}
{\bf Results with SVHN dataset.}
The results for the SVHN dataset are shown in \Cref{table:results}. 
We make the following observations from the results: 
(1) Firstly, MNG-AC significantly outperforms Adv$_{\rm{avg}}$, Adv$_{\rm{max}}$ by $\sim 67.6\%$ and $\sim 18.8\%$ on the Acc$^{\text{union}}_{\text{adv}}$ metric respectively. Furthermore, it achieves an absolute improvement of $+12\%$ on the Acc$^{\text{avg}}_{\text{adv}}$ metric over the multi-perturbation adversarial training baselines. (2) Additionally, we note that compared to MNG-AC, ANT does not improve the performance across adversarial perturbations as it does not utilize them during training.
(3) Interestingly, MNG-AC achieves better performance over the standard $\ell_2$ training with comparable training time, which implies that our method leverages the knowledge across multiple perturbations, illustrating the utility of our method over standard adversarial training.

{\bf Results with Tiny-ImageNet dataset.} We also evaluate our method on Tiny-ImageNet in~\Cref{table:results} to verify that it performs well on complex datasets. We observe that MNG-AC outperforms the multi-perturbation training baselines and achieves comparable performance to the single-perturbation baselines. Only against $\ell_\infty$ perturbations, we notice that Adv$_{\rm{max}}$ achieves marginally better performance.  We believe this is an artefact of the inherent trade-off across multiple perturbations~\citep{tramer2019adversarial, schott2018towards}.
Interestingly, MNG-AC even achieves comparable performance to the single perturbation baselines trained on $\ell_1$ and $\ell_2$ norm. This demonstrates the effectiveness of MNG in preventing overfitting over a single attack, and it's generalization ability to diverse types of attacks.   

{\bf Results with semi-supervised learning.} The efficiency of MNG-AC allows us to utilize semi-supervised data augmentation techniques~\cite{carmon19unlabeled, alayrac19neurips} for multi-perturbation adversarial training with a marginal increase in computation. 
In~\Cref{table:rst}, we can observe that Robust-Self Training (RST$_{\infty}$)~\cite{carmon19unlabeled} overfits to $\ell_\infty$-norm perturbations, while MNG-AC + RST leads to a significant absolute gain of $+11.5\%$ and $+40.3\%$ on the Acc$^{\text{union}}_{\text{adv}}$ metric on CIFAR-10 and SVHN respectively. Furthermore, MNG-AC + RST also improves the absolute performance on the Acc$^{\text{avg}}_{\text{adv}}$ metric by $+14.8\%$ and $+34\%$ on CIFAR-10 and SVHN respectively with comparable training time compared to the RST$_{\infty}$ training.

\begin{table*}[t]
\begin{minipage}{\linewidth}
\begin{minipage}[t]{0.52\textwidth}
\centering
	\caption{Spatial attack evaluation. \label{tab:spatial}}
\resizebox{0.95\linewidth}{!}{
\begin{tabular}{lcccc}
 \toprule
    \multicolumn{1}{c}{Model}&\multicolumn{2}{c}{\textbf{CIFAR-10}} & \multicolumn{2}{c}{\textbf{SVHN}}\\
\midrule
& Accuracy & Time (h) & Accuracy & Time (h)\\
    \midrule
    Adv$_{\rm{avg}}$ & {\bf 54.2{ $\pm$ 0.2}} & 16.9 & 58.7{ $\pm$ 1.8} & 24.1\\ 
    Adv$_{\rm{max}}$ & {\bf 54.3{ $\pm$ 0.2}} & 16.3 & 54.4{ $\pm$ 0.4} & 22.7 \\
    MSD & 52.5{ $\pm$ 0.8} & 16.7 & 45.9{ $\pm$ 0.8} & 23.7\\
    MNG-AC & {52.4{ $\pm$ 0.4}} & 8.4 & {\bf 65.0{ $\pm$ 0.7}} & 11.9\\
    \midrule
    MNG-AC (seen) & \color{red}{\bf 68.9{ $\pm$ 0.8}} & \color{red}{\bf 6.0} & \color{red}{\bf 83.0{ $\pm$ 1.1}} & \color{red}{\bf 10.8} \\
    \bottomrule
\end{tabular}}
		\resizebox{0.322\textwidth}{!}{%
				\begin{tikzpicture}
				\begin{axis}[
	        	title={CIFAR-10},
				xlabel={\HUGE Effect of \scalebox{1.8}{$\beta$}},
				ylabel={\HUGE Adv. acc. (\%)},
				ylabel near ticks,
				xlabel near ticks,
			     width  = 14cm,
			    label style = {font=\Huge},
			    tick label style = {font=\Huge},
                height = 12.3cm,
				xmin=0, xmax=12,
				ymin=36, ymax=42,
				enlarge x limits=true,
				xtick={0,4,8,12},
				legend pos=south east,
				ymajorgrids=true,
				grid=both,
                grid style={line width=.1pt, draw=gray!10},
                major grid style={line width=.2pt,draw=gray!50},
				tickwidth=0.1cm,
				max space between ticks=250,
				font=\HUGE,
				style={ultra thick}
				]
				\addplot[
				color=azure,
				mark=square,
				line width=3.8pt
				]
				coordinates {
					(1,36.7)(4,39.1)(8,40.0)(12,41.4)
				};
				\end{axis}
				\end{tikzpicture}}
		\resizebox{0.29\textwidth}{!}{%
				\begin{tikzpicture}
				\begin{axis}[
	        	title={{SVHN}},
				xlabel={\HUGE Effect of \scalebox{1.5}{$\beta$}},
				 width  = 12cm,
                height = 10.5cm,
				ylabel near ticks,
				xlabel near ticks,
				xmin=1, xmax=12,
				enlarge x limits=true,
				xtick={0,4,8,12},
				legend pos=south east,
				ymajorgrids=true,
				grid=both,
                grid style={line width=.1pt, draw=gray!10},
                major grid style={line width=.2pt,draw=gray!50},
				tickwidth=0.1cm,
				max space between ticks=250,
				font=\HUGE,
				style={ultra thick}
				]
				\addplot[
				color=azure,
				mark=square,
				line width=3.8pt
				]
				coordinates {
					(1,29.3)(4,30.5)(8,31.6)(12,34.2)
				};
				\end{axis}
				\end{tikzpicture}}
		\resizebox{0.29\textwidth}{!}{%
				\begin{tikzpicture}
				\begin{axis}[
	        	title={Tiny-ImageNet},
				xlabel={\HUGE Effect of \scalebox{1.5}{$\beta$}},
				ylabel near ticks,
				 width  = 12cm,
                height = 10.5cm,
				xlabel near ticks,
				xmin=1, xmax=12,
				enlarge x limits=true,
				xtick={0,4,8,12},
				legend pos=south east,
				ymajorgrids=true,
				grid=both,
                grid style={line width=.1pt, draw=gray!10},
                major grid style={line width=.2pt,draw=gray!50},
				tickwidth=0.1cm,
				max space between ticks=250,
				font=\HUGE,
				style={ultra thick}
				]
				\addplot[
				color=azure,
				mark=square,
				line width=3.8pt
				]
				coordinates {
					(1,24.8)(4,27.2)(8,27.6)(12,28.1)
				};
				\end{axis}
				\end{tikzpicture}}
		
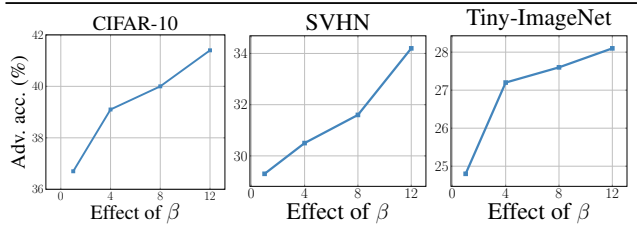
\captionof{figure}{Ablation study on the impact of $\calL_{\rm{ac}}$ on union robustness (Acc$^{\rm{union}}_{\rm{adv}}$) against $\ell_p$ attacks on various datasets. With an increase in $\beta$ in Eq.~\ref{eq:total_loss}, the robustness against the adversarial attacks increases across all the datasets.\label{fig:ablation_plots}}
		\vspace{0.1in}
\end{minipage}\quad
\begin{minipage}[t]{0.45\linewidth}
\begin{figure}[H]
    \centering
    \def\svgwidth{\textwidth}
    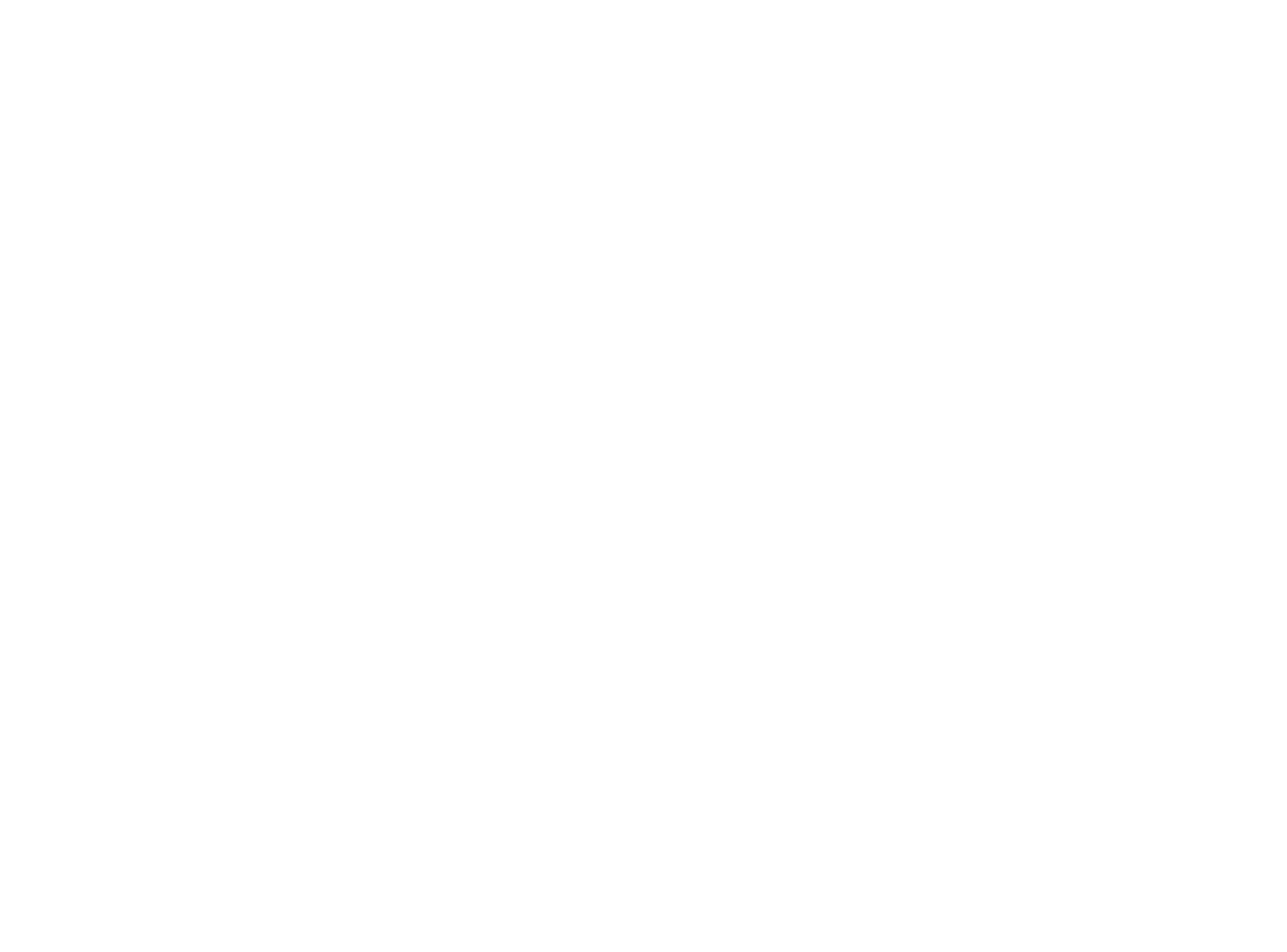

    \caption{Visualization of the loss landscapes for the $\ell_1, \ell_2$, and $\ell_\infty$-norm attacks on the CIFAR-10 dataset. The rows represent the attacks and columns represent different defenses. 
    \label{fig:loss_plots}}
\end{figure}
\end{minipage}
\end{minipage}
\centering
\begin{minipage}[t]{0.23\textwidth}
\fbox{\includegraphics[width=\linewidth]{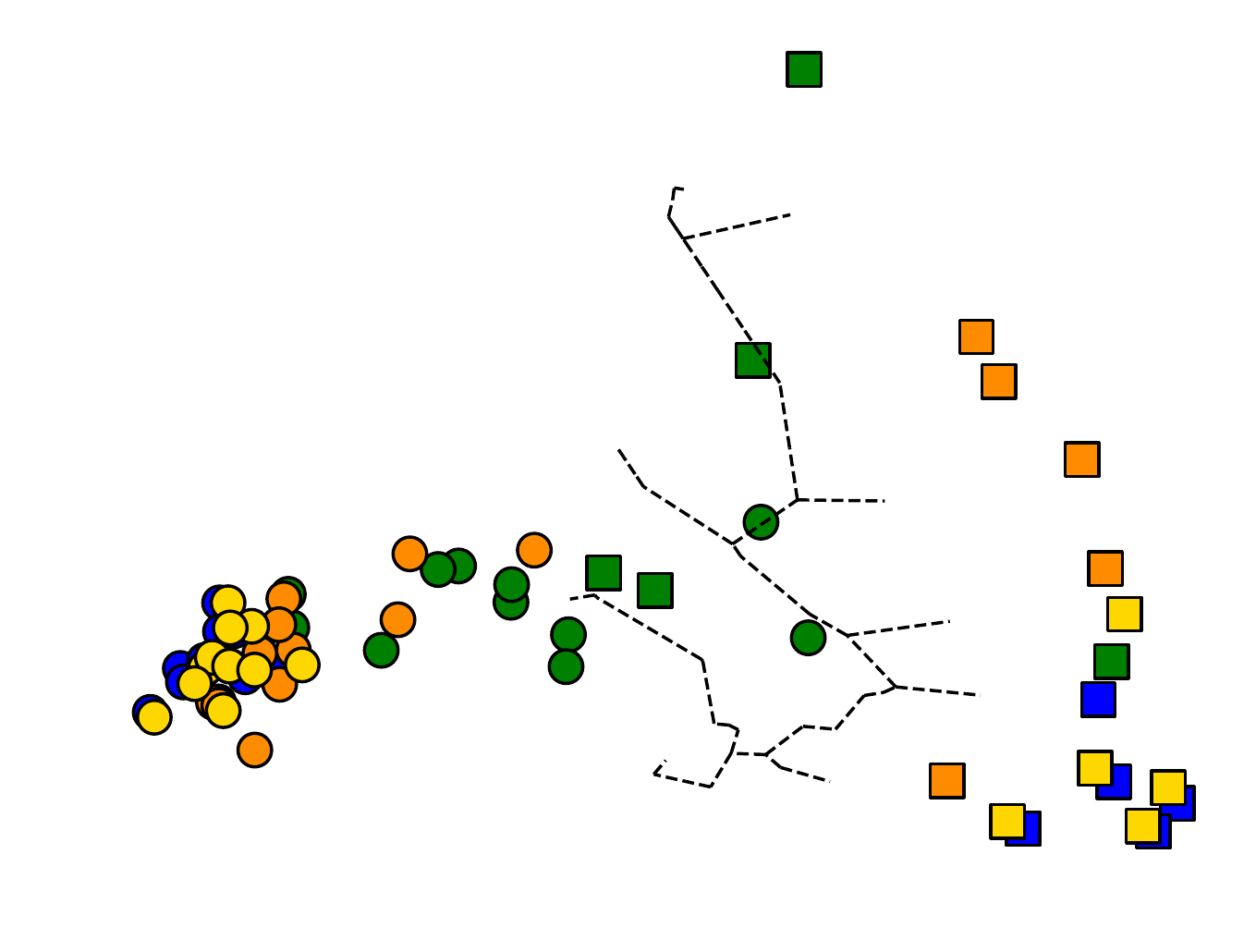}}
\subcaption{Adv$_{\rm avg}$}
\end{minipage}\hfill
\begin{minipage}[t]{0.233\textwidth}
\fbox{\includegraphics[width=\linewidth]{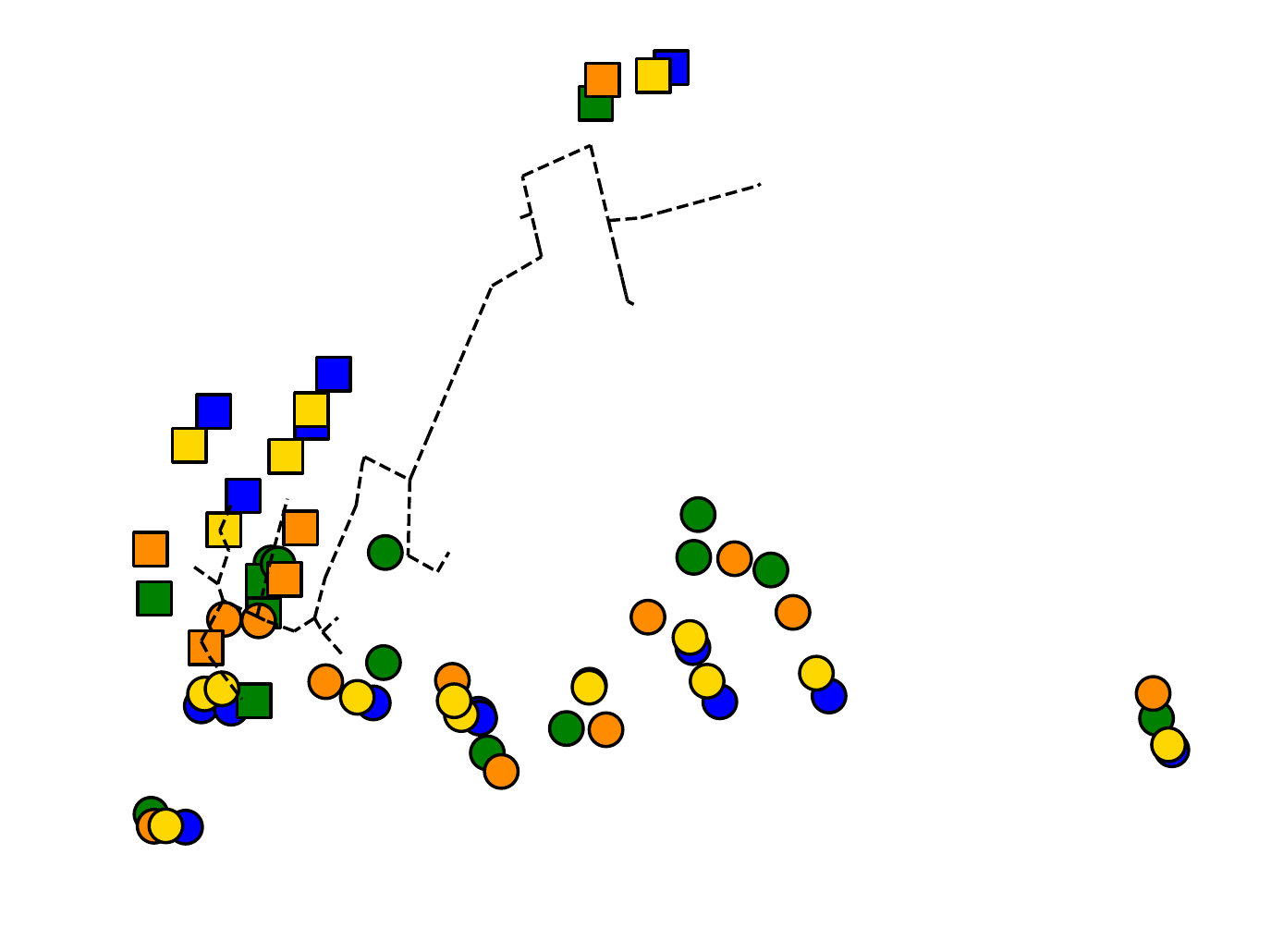}}
\subcaption{Adv$_{\rm max}$}
\end{minipage}\hfill
\begin{minipage}[t]{0.23\textwidth}
\fbox{\includegraphics[width=\linewidth]{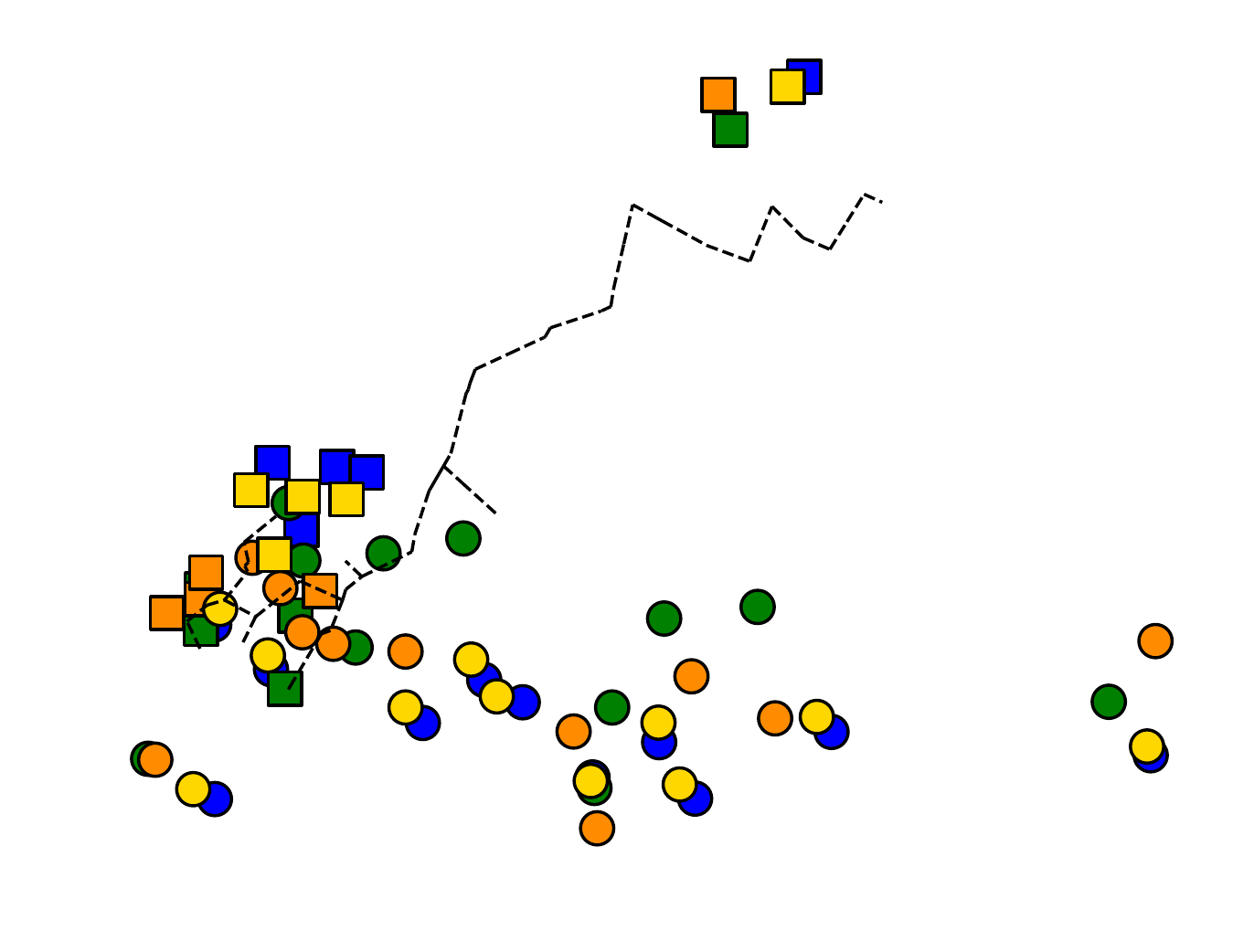}}
\subcaption{MSD}
\end{minipage}\hfill
\begin{minipage}[t]{0.23\textwidth}
\fbox{\includegraphics[width=\linewidth]{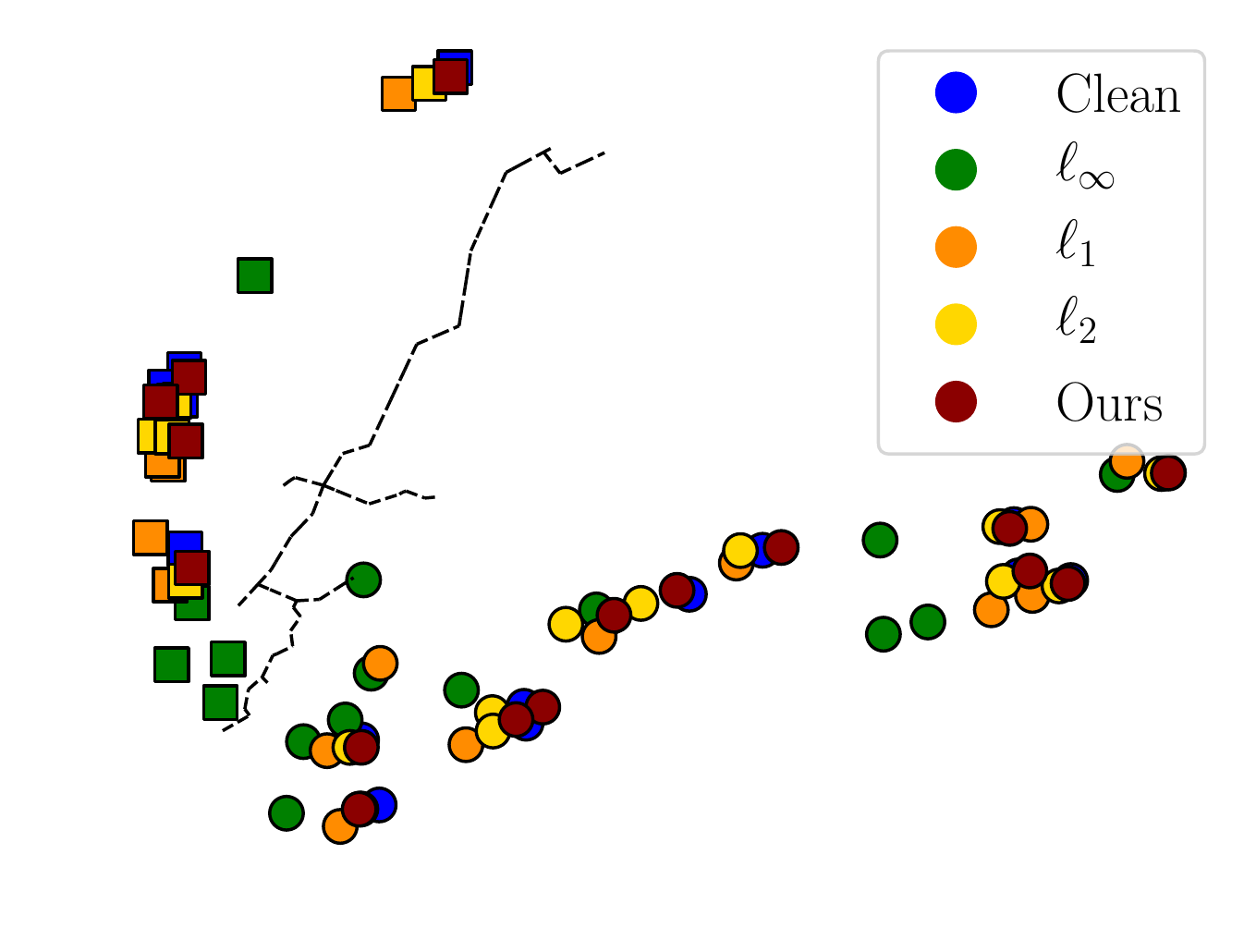}}
\subcaption{MNG-AC}
\end{minipage}\hfill
\captionof{figure}{Visualization of the decision boundary in the penultimate latent-feature space for multi-perturbation methods for SVHN dataset on Wide ResNet 28-10 architecture. The two shapes represent different classes in binary classification.\label{fig:decision_boundary}}
\label{fig:figure3}
\end{table*}
\subsection{Ablation studies}
{\bf Component analysis.} \Cref{tab:ablation_table} dissects the effectiveness of various components in MNG-AC. First, we examine that SAT leads to a $\sim 68\%$ and $\sim 35\%$ relative reduction in training time over multiple perturbations baselines and MNG-AC for both the datasets; however, it does not improve the adversarial robustness. Then, we analyze the impact of our meta-noise generator by injecting random noise $z\sim\calN(0,\bf{I})$ to the inputs for the generation of augmented samples. We observe that it significantly improves the performance over SAT with a marginal increase in the training time. Furthermore, leveraging MNG, our combined framework MNG-AC achieves consistent improvements over all the baselines, demonstrating the efficacy of our meta-learning scheme to defend against multiple perturbations.

{\bf Effect of hyperparameters.} We further analyze the impact of $\beta$ in our augmentation loss (see Eq.~\eqref{eq:total_loss}) in~\Cref{fig:ablation_plots}. In particular, we evaluate the robustness on the Acc$^{\rm{union}}_{\rm{adv}}$ metric across all the $\ell_p$ norm adversarial perturbations for various datasets on Wide ResNet 28-10 architecture. Our results show that as the value of $\beta$ increases the performance improves on the Acc$^{\rm{union}}_{\rm{adv}}$ metric across all the datasets.  Specifically, the absolute performance of Acc$^{\rm{union}}_{\rm{adv}}$ improves by~$\sim5\%$ on all the datasets with an increase in the weight of adversarial consistency loss, highlighting the efficiency of our AC loss. However, we observe that increasing the weight of the consistency loss decreases the clean accuracy by $~\sim3\%$ for all the datasets (see \Cref{fig:appendix_ablation_plots}). We believe that the drop in the clean accuracy is an outcome of the trade-off between clean accuracy and robustness as observed in previous works~\citep{tsipras2018robustness, Zhang2019TheoreticallyPT}, which also holds for adversarial training with multiple perturbations and exploring ways to reduce this trade-off would be an interesting direction for future work.

\subsection{Further analysis of our defense}
{\bf Results with spatial attack.}
We further validated the flexibility and effectiveness of MNG-AC on unseen spatial attacks~\cite{icml19spatial} in \Cref{tab:spatial}. We oberve that MNG-AC largely outperforms the multi-perturbation baselines on the SVHN dataset, and obtains comparable performance to them on CIFAR-10, with significantly smaller training cost. Further, MNG-AC achieves $26.1\%$ and $36.6\%$ relative higher robustness on CIFAR-10 and SVHN respectively when trained jointly with $\ell_p$-norms and spatial attack projected on $\ell_2$-norm ({MNG-AC (seen)}), which is not feasible for baselines due the prohibitive training cost. Moreover, we want to emphasize that MNG-AC (seen) leads to similar performance on $\ell_p$-norms on both the datasets with lower computational cost. Additionally, we evaluate the effectiveness of MNG-AC on CIFAR10-C~\cite{hendrycks2018benchmarking} across five severity levels and unforeseen perturbations~\citep{kang2019testing} (Elastic, $\ell_\infty$-JPEG, $\ell_1$-JPEG and $\ell_2$-JPEG attacks) on SVHN dataset in the \Cref{section:appendix_experiments}.

{\bf Visualization of loss landscape.} 
As further qualitative analysis of the effect of MNG-AC, we compare the loss surface of various methods against $\ell_\infty, \ell_1$, and $\ell_2$ norm attack in~\Cref{fig:loss_plots}. We vary the input along a linear space defined by the $\ell_p$-norm of the gradient where x and y-axes represent the perturbation added in each direction, and the z-axis represents the loss.
We can observe that in most of the instances when trained with a single adversary, the adversary can find a direction orthogonal to that explored during training; for example, $\ell_1$ attack results in a non-smooth loss surface for both $\ell_\infty$ and $\ell_2$ adversarial training. On the contrary, MNG-AC achieves smoother loss surface across all types of attacks which suggests that the gradients modelled by our model are closer to the optimum global landscape. See \Cref{fig:appendix_loss_plots} in \Cref{section:appendix_experiments} for the the loss landscape on multiple $\ell_p$-norm attacks for SVHN dataset.

{\bf Visualization of decision boundary.} We visualize the learned decision boundary in the penultimate latent-feature space on binary-classification task across multiple $\ell_p$-norm attacks in~\Cref{fig:decision_boundary}. In particular, we use use TSNE embedding for obtaining the penultimate latent-features followed by visualizing the decision boundary using the open-source \href{https://github.com/tmadl/highdimensional-decision-boundary-plot}{DBPlot} library. We observe that MNG-AC obtains the least error against all the attacks compared to the baselines trained on multiple adversarial perturbations. Furthermore, note that the adversarial consistency regularization embeds multiple perturbations onto the same latent space, which pushes them away from the decision boundary that in turn improves the overall robustness. Additionally, \Cref{fig:noise} in \Cref{section:appendix_experiments} provides the visualization of the examples generated by our input-dependent meta-noise generator for CIFAR-10 and SVHN dataset.

\section{Conclusion}\label{section:conclusion}
We tackled the problem of robustness against multiple adversarial perturbations. Existing defense methods are tailored to defend against single adversarial perturbation which is an artificial setting to evaluate in real-life scenarios where the adversary will attack the system in any way possible. To this end, we propose a novel \emph{Meta-Noise Generator (MNG)} that learns to stochastically perturb the clean examples by generating output noise across diverse perturbations. Then we train the model using \emph{Adversarial Consistency (AC)} loss that accounts for label consistency across clean, adversarial, and augmented samples. Additionally, to resolve the problem of computation overhead with conventional adversarial training methods for multiple perturbations, we introduce a \emph{Stochastic Adversarial Training (SAT)} which samples a perturbation from the distribution of perturbations. 
We believe that our method can be a strong guideline when other researchers pursue similar tasks in the future.

\section*{Acknowledgements}
We thank the anonymous reviewers for their insightful comments and suggestions. This work was supported by Institute of Information \& communications Technology Planning \& Evaluation (IITP) grant funded by the Korea government (MSIT) (No.2020-0-00153), Penetration Security Testing of ML Model Vulnerabilities and Defense), Engineering Research Center Program through the National Research Foundation of Korea (NRF) funded by the Korean Government MSIT (NRF-2018R1A5A1059921), Institute of Information \& communications Technology Planning \& Evaluation (IITP) grant funded by the Korea government (MSIT)  
(No.2019-0-00075), and Artificial Intelligence Graduate School Program (KAIST). Any opinions, findings, and conclusions or recommendations expressed in this material are those of the authors and do not necessarily reflect the views of the funding agencies.
\bibliography{0_main}
\bibliographystyle{styles/icml2020}

\clearpage
\appendix
{\Large \bf Appendix}

{\bf Organization.} The Appendix is organized as follows: In \Cref{section:appendix_setup}, we describe the hyperparameters and provide the description about evaluation for multiple perturbations. In \Cref{section:appendix_experiments}, we provide additional experiments including the evaluation on common corruptions and unforseen perturbations, effect of the adversarial consistency regularization on clean accuracy, and an expanded comparison of the baselines with our proposed framework on multiple adversarial perturbations. Furthermore, \Cref{section:appendix_experiments} demonstrates the examples generated by our input-dependent meta-noise generator for multiple datasets and visualization of loss landscape on the SVHN dataset.

\section{Experimental setup}\label{section:appendix_setup}

\subsection{Training setup}
We use the SGD optimizer with momentum $0.9$ and weight decay $5\cdot10^{-4}$ to train all our models with cyclic learning rate with a maximum learning rate $\lambda$ that increases linearly from $0$ to $\lambda$ over first $N/2$ epochs and then decreases linearly from $N/2$ to $0$ in the remainder epochs, 
as recommended by~\cite{wong2020fast} for fast convergence of adversarial training. We train all the models with $30$ epochs on a single machine with four GeForce RTX 2080Ti using WideResNet 28-10 architecture for CIFAR-10, SVHN and ResNet-50 for Tiny-ImageNet. We use the maximum learning rate of $\lambda=0.21$ for all our experiments. We use $\beta=12$ (weight for adversarial consistency loss) for the reported results of MNG-AC for all the datasets. 

The noise-generator is formulated as a convolutional network with four 3$\times$3 convolutional layers with LeakyReLU activations and one residual connection from input to output following \cite{rusak2020increasing}. All our algorithms are implemented using Pytorch~\cite{paszke2019pytorch}. We use the weight for the KL divergence ($\beta=6.0$) for TRADES and RST in all our experiments. We replicate all the baselines on SVHN and TinyImageNet since most of the baseline methods have reported their results only on MNIST and CIFAR-10. Moreover, we found that MSD~\cite{maini2019adversarial} and Adv$_{\rm{max}}$ are sensitive to the learning rate on SVHN dataset; therefore, we tune the maximum learning rate and use $\lambda=0.01$ for these baselines. We believe that this is due to the 
the change in optimization formulation, which involves optimization on the worst perturbation and leads to this sensitivity for larger datasets.

\subsection{Evaluation setup}

For CIFAR-10 and SVHN dataset, we use $\varepsilon = \{\frac{8}{255}, \frac{2000}{255}, \frac{128}{255}\}$ and $\alpha=\{0.004, 1.0, 0.1\}$ for $\ell_\infty, \ell_1$, and $\ell_2$ attacks respectively. For Tiny-ImageNet dataset, we use $\varepsilon = \{\frac{4}{255}, \frac{2000}{255}, \frac{80}{255}\}$ and $\alpha=\{0.004, 1.0, 0.1\}$ for $\ell_\infty, \ell_1$, and $\ell_2$ attacks respectively. We use $10$ steps of PGD attack for $\ell_\infty$, $\ell_2$ during training. For $\ell_1$ adversarial training, we use $20$ steps during training and $100$ steps during evaluation. We use the code provided by the authors for evaluation against AutoAttack~\cite{croce2020reliable} and Foolbox~\cite{rauber2017foolbox} library for all the other attacks. 

\section{Additional experimental results}\label{section:appendix_experiments}
\begin{table*}[t!]
\centering
\caption{Average test error (\%) of different corruptions on CIFAR-10 dataset on WideResNet 28-10 over 5 levels of severities of common corruptions. We report the results averaged across 3 runs and five levels of severity for each corruption. \label{tab:common}
}
 \setlength{\tabcolsep}{2pt}
    \renewcommand{\arraystretch}{1.2}
\begin{tabular}{@{}l c | c c c | c c c c | c c c  c | c c c c@{}}
\toprule
 \multicolumn{2}{c|}{} & \multicolumn{3}{c|}{Noise} & \multicolumn{4}{c|}{Blur} & \multicolumn{4}{c|}{Weather} & \multicolumn{4}{c}{Digital} \\
\midrule
Model & All & \scriptsize{Gauss.}
    & \scriptsize{Shot} & \scriptsize{Impulse} & \scriptsize{Defocus} & \scriptsize{Glass} & \scriptsize{Motion} & \scriptsize{Zoom} & \scriptsize{Snow} & \scriptsize{Frost} & \scriptsize{Fog} & \scriptsize{Bright} & \scriptsize{Contrast} & \scriptsize{Elastic} & \scriptsize{Pixel} & \scriptsize{JPEG}\\ 
    \midrule
Adv$_\infty$    & 22.7 & 17.1 & 16.3 & 23.8 & 18.8 & 20.0 & 22.8 & 19.4 & 19.5 & 22.9 & 38.1 & 16.5 & 54.1 &  19.8 & 16.1 & 16.1 \\
Adv$_1$ & 16.6 & 20.0 & 17.3  & 18.1 & 16.0 & 18.5 & 20.7 & 17.9 & 13.4 & 12.6 & 17.7 & 8.2 & 27.6 & 15.0 &  12.7 & 12.3\\
Adv$_2$ & 18.7 & 13.5 & 12.8 & 16.7 & 15.7 & 16.9 & 19.0 & 16.7 & 17.0 & 17.6 & 34.8 & 12.6 & 45.0 & 16.7 & 12.9 & 12.7\\
\midrule
Adv$_{\text{avg}}$  & 21.8 & 16.5 & 15.8 & 16.6 & 18.8 & 19.0 & 21.9 & 19.8 & 20.3 & 22.0 & 38.5 & 16.2 & 49.5 & 19.9 & 16.2 & 15.6 \\
Adv$_{\text{max}}$  & 23.8 & 18.2 & 17.5 & 18.3 & 20.4 & 21.0 & 23.5 & 21.4 & 22.0 & 25.0 & 39.5 & 18.6 & 53.0 & 21.6 & 18.2 & 17.6 \\
MSD & 25.1 & 19.9 & 19.1 & 19.9 & 21.4 & 22.1 & 24.4 & 22.3 & 23.7 & 27.3 & 40.0 & 20.5 & 53.9 & 22.8 & 19.4 & 19.0 \\
\midrule
MNG-AC & 16.6 & 15.3 & 13.8 & 16.4 & 13.2 & 22.0 & 17.0 & 14.5 & 16.2 & 17.3 & 18.6 & 12.7 & 27.1 & 15.8 & 15.7 & 14.2\\
\bottomrule
\end{tabular}
\label{tab:relarchi}
\end{table*}
\begin{table*}[t]
\begin{minipage}{\linewidth}
\centering
\begin{minipage}[t!]{0.5\textwidth}
		\resizebox{0.322\textwidth}{!}{%
				\begin{tikzpicture}
				\begin{axis}[
	        	title={CIFAR-10},
				xlabel={\HUGE Effect of \scalebox{1.8}{$\beta$}},
				ylabel={\HUGE Clean acc. (\%)},
				ylabel near ticks,
				xlabel near ticks,
			     width  = 14cm,
			    label style = {font=\Huge},
			    tick label style = {font=\Huge},
                height = 12.3cm,
				xmin=1, xmax=12,
				enlarge x limits=true,
				xtick={0,4,8,12},
				legend pos=south east,
				ymajorgrids=true,
				grid=both,
                grid style={line width=.1pt, draw=gray!10},
                major grid style={line width=.2pt,draw=gray!50},
				tickwidth=0.1cm,
				max space between ticks=250,
				font=\HUGE,
				style={ultra thick}
				]
				\addplot[
				color=azure,
				mark=square,
				line width=3.8pt
				]
				coordinates {
					(1,86.86)(4,84.97)(8,84.03)(12,81.7)
				};
				\end{axis}
				\end{tikzpicture}}
		\resizebox{0.31\textwidth}{!}{%
				\begin{tikzpicture}
				\begin{axis}[
	        	title={{SVHN}},
				xlabel={\HUGE Effect of \scalebox{1.5}{$\beta$}},
				 width  = 12cm,
                height = 10.5cm,
				ylabel near ticks,
				xlabel near ticks,
				xmin=1, xmax=12,
				enlarge x limits=true,
				xtick={0,4,8,12},
				legend pos=south east,
				ymajorgrids=true,
				grid=both,
                grid style={line width=.1pt, draw=gray!10},
                major grid style={line width=.2pt,draw=gray!50},
				tickwidth=0.1cm,
				max space between ticks=250,
				font=\HUGE,
				style={ultra thick}
				]
				\addplot[
				color=azure,
				mark=square,
				line width=3.8pt
				]
				coordinates {
					(1,93.37)(4,93.27)(8,93.08)(12,92.6)
				};
				\end{axis}
				\end{tikzpicture}}
		\resizebox{0.29\textwidth}{!}{%
				\begin{tikzpicture}
				\begin{axis}[
	        	title={Tiny-ImageNet},
				xlabel={\HUGE Effect of \scalebox{1.5}{$\beta$}},
				ylabel near ticks,
				 width  = 12cm,
                height = 10.5cm,
				xlabel near ticks,
				xmin=1, xmax=12,
				enlarge x limits=true,
				xtick={0,4,8,12},
				legend pos=south east,
				ymajorgrids=true,
				grid=both,
                grid style={line width=.1pt, draw=gray!10},
                major grid style={line width=.2pt,draw=gray!50},
				tickwidth=0.1cm,
				max space between ticks=250,
				font=\HUGE,
				style={ultra thick}
				]
				\addplot[
				color=azure,
				mark=square,
				line width=3.8pt
				]
				coordinates {
					(1,56.63)(4,55.31)(8,54.47)(12, 53.1)
				};
				\end{axis}
				\end{tikzpicture}}
		\captionof{figure}{Ablation study on the impact of $\calL_{\rm{ac}}$ on clean accuracy on various datasets. With an increase in $\beta$, the clean accuracy decreases, due to the inherent accuracy-robustness on all the datasets.\label{fig:appendix_ablation_plots}}
    \end{minipage}%
\begin{minipage}[t!]{0.5\linewidth}
\vspace{-0.8in}
	\caption{Performance of MNG-AC against unforseen adversaries on SVHN dataset.\label{tab:unforseen_table}}
	\resizebox{\linewidth}{!}{
\begin{tabular}{@{}llllll@{}}
\toprule
	Model & Elastic & $\ell_\infty$-JPEG & $\ell_1$-JPEG & $\ell_2$-JPEG  & Acc$^{\rm{union}}_{\rm{adv}}$ \\
	\midrule
	 Adv$_{\text{avg}}$ & 77.1{\scriptsize $\pm$ 1.1} & 86.6{\scriptsize $\pm$ 0.28} & 81.5{\scriptsize $\pm$ 2.1} & 81.2{\scriptsize $\pm$ 1.7} & {62.1 \scriptsize $\pm$ 0.5}\\
	 Adv$_{\text{max}}$ & 60.2{\scriptsize $\pm$ 2.3} & {\bf 89.9{\scriptsize $\pm$ 1.9}} & {\bf 87.9{\scriptsize $\pm$ 2.1}} & {\bf 87.0{\scriptsize $\pm$ 2.5}} & 58.5{\scriptsize $\pm$ 1.5}\\
	 MNG-AC & {\bf 79.6{\scriptsize $\pm$ 1.7}} & 87.5{\scriptsize $\pm$ 0.9} & 75.9{\scriptsize $\pm$ 0.2} & 81.4{\scriptsize $\pm$ 1.2} & {\bf 64.3{\scriptsize $\pm$ 0.5}}\\
	\bottomrule
	\end{tabular}}
	\end{minipage}
\end{minipage}
\end{table*}

{\bf Robustness against common corruptions.} Our sampling strategy further allows us to increase our perturbation set, which is limited in previous works due the increased computation cost. Consequently, we evaluate our method on common corruptions perturbation set~\cite{hendrycks2018benchmarking}. In particular, we use the validation corruptions provided by the authors (speckle noise, gaussian blur, spatter, and saturate) during training and evaluate on other 15 types of unseen corruptions across five levels of severity. We show the results in \Cref{tab:common}. Note that, while the max and multi-steep descent strategies lead to an increase in the test error, MNG-AC achieves significantly better performance compared to the other multi-perturbation baselines. This demonstrates demonstrate the simplicity and effectiveness of MNG-AC on diverese perturbation sets.

{\bf Effect of $\beta$ on clean accuracy.} We further show the effect of $\beta$ on the clean accuracy in \Cref{fig:ablation_plots}. Interestingly, while increasing $\beta$ improves the robustness against multiple adversarial perturbations, it decreases the clean accuracy. In particular, the absolute performance of Acc$^{\rm{union}}_{\rm{adv}}$ improves by~$\sim5\%$, and the clean accuracy drops by $~\sim3\%$ with an increase in the weight of adversarial consistency loss for all the datasets. We report the MNG-AC results with $\beta=12$ for all our experiments to achieve an optimal trade-off for our proposed method. 

{\bf Expanded results.}  Due to the length limit of our paper, we provide a breakdown of all the attacks on CIFAR-10 in \Cref{tab:cifar_appendix}, SVHN on Wide ResNet 28-10 in \Cref{tab:svhn_appendix}, Tiny-ImageNet on ResNet50 in \Cref{tab:tiny_appendix}. 

{\bf Results on unforseen adversaries.}
We further evaluate our model on various unforeseen perturbations~\citep{kang2019testing} namely we evaluate on the Elastic, $\ell_\infty$-JPEG, $\ell_1$-JPEG and $\ell_2$-JPEG attacks. Note that, even though adversarial training methods do not generalize beyond the threat model, we observe that MNG-AC improves the performance on these unseen adversaries. We compare our MNG-AC to the baselines trained with multiple perturbations on the SVHN dataset in \Cref{tab:unforseen_table}. We notice that even though, Adv$_{\rm{max}}$ achieves better performance on $\ell_p$-JPEG attacks, it obtains the minimum robustness across the Acc$^{\text{union}}_{\text{adv}}$ metric. In contrast, MNG-AC generalizes better over both the baselines for the worst-attack and shows a relative gain of $+3.5\%$
over the best performing baseline.

{\bf Visualization of generated examples.}
We visualize the generated examples by our generator during training by randomly selecting samples projected on various $\ell_p$ norms and datasets in \Cref{fig:noise}. From the figure, we can observe that our meta-noise generator incorporates the features by different attacks and learns diverse input-dependent noise distributions across multiple adversarial perturbations by explicitly minimizing the adversarial loss across multiple perturbations during meta-training. Overall, it combines two complementary approaches and leads to a novel input-dependent learner for generalization across diverse attacks.
\begin{table*}[t!]
\vspace{-0.18em}%
  \caption{Summary of adversarial accuracy results for CIFAR-10 on Wide ResNet 28-10 architecture. The best and second-best results are highlighted in bold and underline respectively.\label{tab:cifar_appendix}}
    \centering
    \footnotesize
    \setlength{\tabcolsep}{2pt}
    \renewcommand{\arraystretch}{1.2}
  \begin{tabular}{lcccccccccl}

    \toprule
    & Adv$_\infty$ & Adv$_1$ & Adv$_2$ & Trades$_\infty$ & RST$_\infty$ & Adv$_{\text{avg}}$ & Adv$_{\text{max}}$ & MSD & MNG-AC & MNG-AC + RST\\
                                             
     \cmidrule{2-11}
Clean Accuracy  & 86.8{\scriptsize $\pm$ 0.1} & 93.3{\scriptsize $\pm$ 0.6} & 91.7{\scriptsize $\pm$ 0.2} & 84.7{\scriptsize $\pm$ 0.3} & 88.9{\scriptsize $\pm$ 0.2} & 87.1{\scriptsize $\pm$ 0.2} & 85.4{\scriptsize $\pm$ 0.3} & 82.3{\scriptsize $\pm$ 0.2} & 84.9{\scriptsize $\pm$ 0.3} & 88.7{\scriptsize $\pm$ 0.2}\\
\midrule

PGD-$\ell_\infty$   & 46.9{\scriptsize $\pm$ 0.5} & 0.40{\scriptsize $\pm$ 0.7} & 30.4{\scriptsize $\pm$ 1.4} & 52.0{\scriptsize $\pm$ 0.6} & 56.9{\scriptsize $\pm$ 0.1} & 35.7{\scriptsize $\pm$ 0.5} & 42.5{\scriptsize $\pm$ 0.4} & 46.3{\scriptsize $\pm$ 0.6} & 45.4{\scriptsize $\pm$ 0.8} & 52.8{\scriptsize $\pm$ 0.9}\\

PGD-Foolbox  & 54.7{\scriptsize $\pm$ 0.4} & 0.33{\scriptsize $\pm$ 0.6} & 40.9{\scriptsize $\pm$ 0.9} & 57.8{\scriptsize $\pm$ 0.5} & 62.9{\scriptsize $\pm$ 0.3} & 45.0{\scriptsize $\pm$ 0.4} & 50.4{\scriptsize $\pm$ 0.4} & 52.9{\scriptsize $\pm$ 0.8} & 52.1{\scriptsize $\pm$ 0.6} & 59.0{\scriptsize $\pm$ 0.7} \\

AutoAttack  & 44.9{\scriptsize $\pm$ 0.7} & 0.0{\scriptsize $\pm$ 0.0} & 28.8{\scriptsize $\pm$ 1.3} & 48.9{\scriptsize $\pm$ 0.9} & 54.9{\scriptsize $\pm$ 0.3} & 34.2{\scriptsize $\pm$ 0.5} & 39.9{\scriptsize $\pm$ 0.5} & 43.5{\scriptsize $\pm$ 0.5} & 41.4{\scriptsize $\pm$ 0.7} & 47.2{\scriptsize $\pm$ 0.8} \\

Brendel \& Bethge & 49.9{\scriptsize $\pm$ 1.1} & 0.0{\scriptsize $\pm$ 0.0} & 35.4{\scriptsize $\pm$ 1.0} & 52.1{\scriptsize $\pm$ 0.7} & 56.5{\scriptsize $\pm$ 1.8} & 40.2{\scriptsize $\pm$ 1.5} & 45.5{\scriptsize $\pm$ 0.9} & 49.3{\scriptsize $\pm$ 1.1} & 47.0{\scriptsize $\pm$ 0.9} & 53.4{\scriptsize $\pm$ 0.8}\\
\cmidrule{2-11}

{\bf All $\ell_\infty$ attacks} & 44.9{\scriptsize $\pm$ 0.7} & {0.0\scriptsize $\pm$ 0.0} & 28.8{\scriptsize $\pm$ 1.3} & 48.9{\scriptsize $\pm$ 0.7} & 54.9{\scriptsize $\pm$ 1.8} & 34.1{\scriptsize $\pm$ 0.5} & 39.9{\scriptsize $\pm$ 0.5} & 43.5{\scriptsize $\pm$ 0.5} & 41.4{\scriptsize $\pm$ 0.7} & 47.2{\scriptsize $\pm$ 0.8}\\
\midrule

PGD-$\ell_1$ & 26.4{\scriptsize $\pm$ 0.5}  & 93.9{\scriptsize $\pm$ 0.6}  & 54.4{\scriptsize $\pm$ 0.6}  & 32.4{\scriptsize $\pm$ 1.0}  & 36.2{\scriptsize $\pm$ 0.6}  & 61.4{\scriptsize $\pm$ 0.6}  & 57.9{\scriptsize $\pm$ 0.6}  & 54.4{\scriptsize $\pm$ 0.7} & 65.4{\scriptsize $\pm$ 0.2} & 73.8{\scriptsize $\pm$ 0.2} \\

PGD-Foolbox  & 35.2{\scriptsize $\pm$ 0.7}  & 92.3{\scriptsize $\pm$ 1.3}  & 54.2{\scriptsize $\pm$ 0.5}  & 40.3{\scriptsize $\pm$ 0.7}  & 44.6{\scriptsize $\pm$ 0.3}  & 64.5{\scriptsize $\pm$ 0.2}  & 60.7{\scriptsize $\pm$ 0.5}  & 60.3{\scriptsize $\pm$ 0.4} & 65.5{\scriptsize $\pm$ 0.1} & 74.9{\scriptsize $\pm$ 0.7}\\

EAD  & 72.9{\scriptsize $\pm$1.0} & 87.1{\scriptsize $\pm$ 3.3}  & 75.9{\scriptsize $\pm$ 1.9} & 80.2{\scriptsize $\pm$ 0.7}  & 84.5{\scriptsize $\pm$ 0.2}  & 85.7{\scriptsize $\pm$ 0.2} & 83.3{\scriptsize $\pm$ 0.5}  & 80.8{\scriptsize $\pm$ 0.1} & 79.3{\scriptsize $\pm$ 0.6} & 88.4{\scriptsize $\pm$ 0.5}\\

SAPA    &  71.5{\scriptsize $\pm$ 0.2}  & 80.7{\scriptsize $\pm$ 1.8}  &  81.9{\scriptsize $\pm$ 0.5}  & 71.4{\scriptsize $\pm$ 0.7}  & 76.0{\scriptsize $\pm$ 0.5}  & 82.7{\scriptsize $\pm$ 0.1}  & 80.0{\scriptsize $\pm$ 0.1}  & 76.9{\scriptsize $\pm$ 0.5} & 76.7{\scriptsize $\pm$ 0.4}& 85.4{\scriptsize $\pm$ 0.3} \\
\cmidrule{2-11}

{\bf All $\ell_1$ attacks} & 26.2{\scriptsize $\pm$ 0.4} & 80.7{\scriptsize $\pm$ 0.7}  & 54.2{\scriptsize $\pm$ 0.4}  & 32.3{\scriptsize $\pm$ 1.0}  & 36.0{\scriptsize $\pm$ 0.9}  & 61.3{\scriptsize $\pm$ 0.6}  & 57.9{\scriptsize $\pm$ 0.7}  & 54.3{\scriptsize $\pm$ 0.4} & 65.4{\scriptsize $\pm$ 0.3} & 73.8{\scriptsize $\pm$ 0.7} \\
\midrule

PGD-$\ell_2$ & 57.1{\scriptsize $\pm$ 0.4} & 3.0{\scriptsize $\pm$ 0.9} & 66.2{\scriptsize $\pm$ 0.2} &  60.8{\scriptsize $\pm$ 0.8} & 62.4{\scriptsize $\pm$ 0.2} & 66.5{\scriptsize $\pm$ 0.4} & 66.4{\scriptsize $\pm$ 0.2} & 65.0{\scriptsize $\pm$ 0.2} & 67.2{\scriptsize $\pm$ 0.2} & 76.7{\scriptsize $\pm$ 0.7}\\

PGD-Foolbox & 65.9{\scriptsize $\pm$ 0.7} & 3.4{\scriptsize $\pm$ 1.9} & 72.0{\scriptsize $\pm$ 0.4} & {66.2\scriptsize $\pm$ 0.6} & {70.8\scriptsize $\pm$ 0.3} & {70.1\scriptsize $\pm$ 0.1} & {69.7\scriptsize $\pm$ 0.7} & 68.6{\scriptsize $\pm$ 0.2} & {70.9\scriptsize $\pm$ 0.3}& 79.0{\scriptsize $\pm$ 0.3}\\

Gaussian Noise & 84.6{\scriptsize $\pm$ 0.5} & 81.0{\scriptsize $\pm$ 2.3} & 88.5{\scriptsize $\pm$ 0.4} & 82.4{\scriptsize $\pm$ 0.6} & 87.4{\scriptsize $\pm$ 0.2} & 84.5{\scriptsize $\pm$ 0.6} & 81.9{\scriptsize $\pm$ 0.4} & 80.7{\scriptsize $\pm$ 0.8} & 79.9{\scriptsize $\pm$ 0.3}& 87.7{\scriptsize $\pm$ 0.4}\\

AutoAttack & 55.1{\scriptsize $\pm$ 0.8} & 0.0{\scriptsize $\pm$ 0.0} & 65.8{\scriptsize $\pm$ 0.3} & 57.8{\scriptsize $\pm$ 0.6} & 59.8{\scriptsize $\pm$ 0.2} & 65.7{\scriptsize $\pm$ 0.4} & 64.5{\scriptsize $\pm$ 0.1} & 63.1{\scriptsize $\pm$ 0.5} & 65.2{\scriptsize $\pm$ 0.5}& 73.7{\scriptsize $\pm$ 0.2}\\

Brendel \& Bethge  & 59.6{\scriptsize $\pm$ 1.2} & 0.0{\scriptsize $\pm$ 0.0} & 68.6{\scriptsize $\pm$ 0.2} & 60.9{\scriptsize $\pm$ 1.0} & 62.9{\scriptsize $\pm$ 0.7} & 67.5{\scriptsize $\pm$ 0.2} & 66.6{\scriptsize $\pm$ 0.2} & 66.4{\scriptsize $\pm$ 0.3} & 66.6{\scriptsize $\pm$ 0.7} & 75.3{\scriptsize $\pm$ 0.2} \\

CWL2 & 57.5{\scriptsize $\pm$ 0.9} & 0.1{\scriptsize $\pm$ 0.0} & 66.7{\scriptsize $\pm$ 0.3} & 59.3{\scriptsize $\pm$ 0.4} & 60.9{\scriptsize $\pm$ 0.3} & 66.8{\scriptsize $\pm$ 0.2} & 65.4{\scriptsize $\pm$ 0.3} & 64.1{\scriptsize $\pm$ 0.3} & 66.5{\scriptsize $\pm$ 0.6} & 74.2{\scriptsize $\pm$ 0.5}\\
      \cmidrule{2-11}

{\bf All $\ell_2$ attacks} & 55.0{\scriptsize $\pm$ 0.9} & 0.0{\scriptsize $\pm$ 0.0} & 65.8{\scriptsize $\pm$ 0.3} & {57.8\scriptsize $\pm$ 0.6} & {59.5\scriptsize $\pm$ 0.2} & {65.7\scriptsize $\pm$ 0.4} & {64.5\scriptsize $\pm$ 0.1} & 63.1{\scriptsize $\pm$ 0.5} & 65.2{\scriptsize $\pm$ 0.5} & 73.7{\scriptsize $\pm$ 0.2}\\
   \midrule
     Acc$^{\rm{union}}_{\rm{adv}}$ & 25.6{\scriptsize $\pm$ 0.6} & 0.0{\scriptsize $\pm$ 0.0} & 28.6{\scriptsize $\pm$ 1.4} & 31.5{\scriptsize $\pm$ 1.2} & 35.7{\scriptsize $\pm$ 0.6} & 34.1{\scriptsize $\pm$ 0.1} & 39.7{\scriptsize $\pm$ 0.5} & \underline{42.7{\scriptsize $\pm$ 0.5}} & {41.4{\scriptsize $\pm$ 0.7}} & {\bf 47.2{\scriptsize $\pm$ 0.7}}\\
     Acc$^{\rm{avg}}_{\rm{adv}}$ & 41.9{\scriptsize $\pm$ 0.6} & 26.8{\scriptsize $\pm$ 0.6} & 49.6{\scriptsize $\pm$ 0.3} & 46.3{\scriptsize $\pm$ 0.7} & 50.1{\scriptsize $\pm$ 0.8} & 53.7{\scriptsize $\pm$ 0.3} & 54.1{\scriptsize $\pm$ 0.4} & 53.6{\scriptsize $\pm$ 0.2} & \underline{57.2{\scriptsize $\pm$ 0.4}} & {\bf 64.9{\scriptsize $\pm$ 0.3}} \\
    \bottomrule
  \end{tabular}
\end{table*}\noindent
\begin{table*}[h!]
  \caption{Summary of adversarial accuracy results for SVHN dataset on Wide ResNet 28-10 architecture. \label{tab:svhn_appendix}}
    \centering
    \footnotesize
    \setlength{\tabcolsep}{2pt}
    \renewcommand{\arraystretch}{1.2}
  \begin{tabular}{lcccccccccl}
    \toprule
    & Adv$_\infty$ & Adv$_1$ & Adv$_2$ & Trades$_\infty$ & RST$_\infty$ & Adv$_{\text{avg}}$ & Adv$_{\text{max}}$ & MSD & MNG-AC & MNG-AC + RST \\
                                             
     \cmidrule{2-11}
    Clean Accuracy  & 92.8{\scriptsize $\pm$ 0.1} & 92.4{\scriptsize $\pm$ 1.6} & 94.9{\scriptsize $\pm$ 0.0} & 93.9{\scriptsize $\pm$ 0.0} & 95.6{\scriptsize $\pm$ 0.0} & 92.6{\scriptsize $\pm$ 0.1} & 86.9{\scriptsize $\pm$ 0.3} & 81.8{\scriptsize $\pm$ 0.3} & 93.4{\scriptsize $\pm$ 0.0} & 96.3{\scriptsize $\pm$ 0.3}  \\
    \midrule
    
    PGD-$\ell_\infty$  & 49.1{\scriptsize $\pm$ 0.1} & 3.2{\scriptsize $\pm$ 2.4} & 29.3{\scriptsize $\pm$ 0.4} & 55.5{\scriptsize $\pm$ 1.4} & 66.9{\scriptsize $\pm$ 0.8} & 24.9{\scriptsize $\pm$ 2.7} & 32.7{\scriptsize $\pm$ 0.6} & 39.7{\scriptsize $\pm$ 0.7} & 42.6{\scriptsize $\pm$ 0.5} & 58.0{\scriptsize $\pm$ 1.4}\\
    
    PGD-Foolbox  & 60.7{\scriptsize $\pm$ 0.4} & 2.5{\scriptsize $\pm$ 1.9} & 43.2{\scriptsize $\pm$ 1.3} & 66.4{\scriptsize $\pm$ 1.1} & 73.8{\scriptsize $\pm$ 0.3} & 37.1{\scriptsize $\pm$ 3.1} & 45.6{\scriptsize $\pm$ 0.2} & 48.5{\scriptsize $\pm$ 0.2} & 56.1{\scriptsize $\pm$ 0.9} & 66.8{\scriptsize $\pm$ 0.6}\\
    
    AutoAttack   & 46.2{\scriptsize $\pm$ 0.6} & 0.0{\scriptsize $\pm$ 0.0} & 21.8{\scriptsize $\pm$ 0.3} & 49.9{\scriptsize $\pm$ 1.8} & 61.0{\scriptsize $\pm$ 2.0} & 21.5{\scriptsize $\pm$ 2.8} & 28.8{\scriptsize $\pm$ 0.2} & 34.1{\scriptsize $\pm$ 0.6} & 34.2{\scriptsize $\pm$ 1.0} & 43.8{\scriptsize $\pm$ 1.5}\\
    
    Brendel \& Bethge & 51.6{\scriptsize $\pm$ 0.7} & 0.0{\scriptsize $\pm$ 0.0} & 26.5{\scriptsize $\pm$ 0.9} & 55.8{\scriptsize $\pm$ 1.5} & 65.6{\scriptsize $\pm$ 1.2} & 24.5{\scriptsize $\pm$ 2.9} & 36.4{\scriptsize $\pm$ 0.4} & 41.7{\scriptsize $\pm$ 0.2} & 42.1{\scriptsize $\pm$ 1.9} & 50.7{\scriptsize $\pm$ 0.9}\\
    \cmidrule{2-11}
    
    {\bf All $\ell_\infty$ attacks} & 46.2{\scriptsize $\pm$ 0.6} & 0.0{\scriptsize $\pm$ 0.0} & 21.7{\scriptsize $\pm$ 0.4} & 49.9{\scriptsize $\pm$ 1.7} & 60.9{\scriptsize $\pm$ 2.0} & 21.5{\scriptsize $\pm$ 2.7} & 28.8{\scriptsize $\pm$ 0.2} & 34.1{\scriptsize $\pm$ 0.6} & 34.2{\scriptsize $\pm$ 1.0} & 43.8{\scriptsize $\pm$ 1.5}\\
       \midrule
    PGD-$\ell_1$ & 10.0{\scriptsize $\pm$ 0.3} & 97.5{\scriptsize $\pm$ 1.3} & 45.2{\scriptsize $\pm$ 0.3} & 4.8{\scriptsize $\pm$ 0.4} & 3.6{\scriptsize $\pm$ 0.4} & 62.3{\scriptsize $\pm$ 3.9} & 48.9{\scriptsize $\pm$ 0.9} & 43.4{\scriptsize $\pm$ 0.5} & 71.6{\scriptsize $\pm$ 2.0} & 78.9{\scriptsize $\pm$ 2.0}\\
    PGD-Foolbox  & 19.9{\scriptsize $\pm$ 0.8} & 94.6{\scriptsize $\pm$ 0.4} & 57.5{\scriptsize $\pm$ 0.1 } & 15.5{\scriptsize $\pm$ 0.2} & 11.3{\scriptsize $\pm$ 0.5} & 79.2{\scriptsize $\pm$ 3.4} & 52.8{\scriptsize $\pm$ 0.2} & 48.2{\scriptsize $\pm$ 0.2} & 73.3{\scriptsize $\pm$ 0.7} & 82.0{\scriptsize $\pm$ 0.3} \\
    EAD & 65.7{\scriptsize $\pm$ 2.1} & 87.8{\scriptsize $\pm$ 1.9} & 82.3{\scriptsize $\pm$ 1.2} & 51.5{\scriptsize $\pm$ 2.9} & 60.4{\scriptsize $\pm$ 0.8} & 84.8{\scriptsize $\pm$ 2.4} & 85.7{\scriptsize $\pm$ 0.3} & 81.1{\scriptsize $\pm$ 0.2} & 92.1{\scriptsize $\pm$ 2.2} & 95.8{\scriptsize $\pm$ 0.3}\\
    SAPA     & 79.4{\scriptsize $\pm$ 0.8} & 77.3{\scriptsize $\pm$ 5.2} & 87.3{\scriptsize $\pm$ 0.1} & 73.5{\scriptsize $\pm$ 1.0} & 86.2{\scriptsize $\pm$ 0.5} & 88.5{\scriptsize $\pm$ 0.6} & 81.4{\scriptsize $\pm$ 0.2} & 75.6{\scriptsize $\pm$ 0.3} & 89.9{\scriptsize $\pm$ 1.6} & 94.1{\scriptsize $\pm$ 0.2}\\
    \cmidrule{2-11}
    
    {\bf All $\ell_1$ attacks}  & 8.2{\scriptsize $\pm$ 0.9} & 77.2{\scriptsize $\pm$ 2.9} & 44.7{\scriptsize $\pm$ 0.5} & 4.2{\scriptsize $\pm$ 0.4} & 3.5{\scriptsize $\pm$ 0.5} & 61.2{\scriptsize $\pm$ 4.1} & 48.9{\scriptsize $\pm$ 0.9} & 43.4{\scriptsize $\pm$ 0.5} & 71.3{\scriptsize $\pm$ 1.7} & 78.9{\scriptsize $\pm$ 2.0} \\
    \midrule
    
    PGD-$\ell_2$ & 36.3{\scriptsize $\pm$ 0.9} & 3.4{\scriptsize $\pm$ 1.4} & 63.6{\scriptsize $\pm$ 0.5} & 34.4{\scriptsize $\pm$ 2.0} & 35.2{\scriptsize $\pm$ 0.7} & 60.5{\scriptsize $\pm$ 0.2} & 59.2{\scriptsize $\pm$ 0.6} & 56.3{\scriptsize $\pm$ 0.3} & 72.3{\scriptsize $\pm$ 0.3} & 80.1{\scriptsize $\pm$ 0.4}   \\
    
    PGD-Foolbox  & 55.7{\scriptsize $\pm$ 0.1} & 4.2{\scriptsize $\pm$ 1.8} & 72.3{\scriptsize $\pm$ 0.9} & 56.0{\scriptsize $\pm$ 0.2} & 56.7{\scriptsize $\pm$ 1.0} &  70.3{\scriptsize $\pm$ 0.6} & 66.3{\scriptsize $\pm$ 0.4} & 61.9{\scriptsize $\pm$ 0.7} & 77.3{\scriptsize $\pm$ 0.2} & 83.3{\scriptsize $\pm$ 0.2}  \\
    
    Gaussian Noise  & 91.8{\scriptsize $\pm$ 0.1} & 69.3{\scriptsize $\pm$ 2.5} & 91.8{\scriptsize $\pm$ 0.2} & 93.5{\scriptsize $\pm$ 0.3} & 92.5{\scriptsize $\pm$ 0.4} & 90.7{\scriptsize $\pm$ 0.9} & 86.3{\scriptsize $\pm$ 0.5} & 80.6{\scriptsize $\pm$ 0.7} & 92.0{\scriptsize $\pm$ 0.4} & 94.0{\scriptsize $\pm$ 0.5}  \\
    
    AutoAttack & 30.2{\scriptsize $\pm$ 0.5} & 0.0{\scriptsize $\pm$ 0.0} & 62.9{\scriptsize $\pm$ 0.2} & 28.0{\scriptsize $\pm$ 2.2} & 28.9{\scriptsize $\pm$ 0.8} & 58.0{\scriptsize $\pm$ 1.7} & 56.3{\scriptsize $\pm$ 0.8} & 54.1{\scriptsize $\pm$ 0.1} & 66.7{\scriptsize $\pm$ 0.9} & 72.6{\scriptsize $\pm$ 0.1}  \\
    
    Brendel \& Bethge  & 41.8{\scriptsize $\pm$ 0.8} & 0.0{\scriptsize $\pm$ 0.0} & 67.0{\scriptsize $\pm$ 0.9} & 39.9{\scriptsize $\pm$ 1.3} & 47.8{\scriptsize $\pm$ 0.6} & 61.4{\scriptsize $\pm$ 2.3} & 60.9{\scriptsize $\pm$ 0.9} & 57.9{\scriptsize $\pm$ 0.8} & 71.2{\scriptsize $\pm$ 1.0} & 78.0{\scriptsize $\pm$ 0.3} \\
    
    CWL2    & 39.4{\scriptsize $\pm$ 0.3} & 0.0{\scriptsize $\pm$ 0.0} & 54.8{\scriptsize $\pm$ 0.2} & 35.4{\scriptsize $\pm$ 1.9} & 45.0{\scriptsize $\pm$ 0.5} & 61.5{\scriptsize $\pm$ 0.6} & 57.8{\scriptsize $\pm$ 1.2} & 55.2{\scriptsize $\pm$ 0.4} & 69.2{\scriptsize $\pm$ 0.9} & 74.2{\scriptsize $\pm$ 0.5} \\
    
      \cmidrule{2-11}
    {\bf All $\ell_2$ attacks}  & 30.2{\scriptsize $\pm$ 0.5} & 0.0{\scriptsize $\pm$ 0.0} & 62.9{\scriptsize $\pm$ 0.2} & 26.7{\scriptsize $\pm$ 2.0} & 28.8{\scriptsize $\pm$ 0.9} & 56.1{\scriptsize $\pm$ 2.3} & 56.3{\scriptsize $\pm$ 0.8} & 54.1{\scriptsize $\pm$ 0.1} & 66.7{\scriptsize $\pm$ 0.9} & 72.6{\scriptsize $\pm$ 0.2} \\
 
  \midrule
    Acc$^{\rm{union}}_{\rm{adv}}$ &  8.1{\scriptsize $\pm$ 0.9} & 0.0{\scriptsize $\pm$ 0.0} & 21.0{\scriptsize $\pm$ 0.4} & 4.1{\scriptsize $\pm$ 0.4} & 3.5{\scriptsize $\pm$ 0.5} & 20.4{\scriptsize $\pm$ 2.7} & 28.8{\scriptsize $\pm$ 0.2} & 34.1{\scriptsize $\pm$ 0.6} & \underline{34.2{\scriptsize $\pm$ 1.0}} & {\bf 43.8{\scriptsize $\pm$ 1.5}}  \\
    Acc$^{\rm{avg}}_{\rm{adv}}$ &  28.3{\scriptsize $\pm$ 0.1} & 25.7{\scriptsize $\pm$ 1.0} & 43.1{\scriptsize $\pm$ 0.3} & 26.9{\scriptsize $\pm$ 1.1} & 31.1{\scriptsize $\pm$ 0.6} & 45.9{\scriptsize $\pm$ 0.9} & 44.7{\scriptsize $\pm$ 0.4} & 44.0{\scriptsize $\pm$ 0.1} & \underline{ 57.4{\scriptsize $\pm$ 0.4}}  & {\bf 65.1{\scriptsize $\pm$ 0.3}} \\
    \bottomrule
  \end{tabular}
\end{table*}\noindent
\begin{table*}[h!]
  \caption{Summary of adversarial accuracy results for Tiny-ImageNet on ResNet50 architecture.\label{tab:tiny_appendix}}
    \centering
    \footnotesize
    \setlength{\tabcolsep}{2pt}
    \renewcommand{\arraystretch}{1.2}
  \begin{tabular}{lcccccccccc}
    \toprule
    & Adv$_\infty$ & Adv$_1$ & Adv$_2$ & Trades$_\infty$ & Adv$_{\text{avg}}$ & Adv$_{\text{max}}$ & MSD & MNG-AC\\
     \cmidrule{2-9}
    Clean Accuracy  & 54.2{\scriptsize $\pm$ 0.1}  & 57.8{\scriptsize $\pm$ 0.2}  & 59.8{\scriptsize $\pm$ 0.1}  & 48.2{\scriptsize $\pm$ 0.2}  & 56.0{\scriptsize $\pm$ 0.2}  & 53.5{\scriptsize $\pm$ 0.0} & 53.0{\scriptsize $\pm$ 0.1} & 53.1{\scriptsize $\pm$ 0.1}\\
    \midrule
    PGD-$\ell_\infty$  & 32.1{\scriptsize $\pm$ 0.0} & 11.5{\scriptsize $\pm$ 1.2}  & 17.9{\scriptsize $\pm$ 1.1}  & 32.2{\scriptsize $\pm$ 0.4} & 25.0{\scriptsize $\pm$ 0.6} & 32.0{\scriptsize $\pm$ 0.6} & 31.9{\scriptsize $\pm$ 0.4} & 29.3{\scriptsize $\pm$ 0.3} \\
    PGD-Foolbox  & 34.6{\scriptsize $\pm$ 0.4}  & 17.2{\scriptsize $\pm$ 0.1}  & 5.2{\scriptsize $\pm$ 0.6}  & 34.1{\scriptsize $\pm$ 0.2} & 34.0{\scriptsize $\pm$ 0.2}  & 29.8{\scriptsize $\pm$ 0.1}  & 34.4{\scriptsize $\pm$ 0.4} & 32.3{\scriptsize $\pm$ 0.3}  \\
    AutoAttack   & 29.6{\scriptsize $\pm$ 0.1}  & 10.5{\scriptsize $\pm$ 0.7}  & 16.3{\scriptsize $\pm$ 0.3}  & 28.7{\scriptsize $\pm$ 0.9}  &  23.7{\scriptsize $\pm$ 0.2}  & 30.0{\scriptsize $\pm$ 0.1} & 29.4{\scriptsize $\pm$ 0.3} & 28.1{\scriptsize $\pm$ 0.4}\\
    Brendel \& Bethge & 32.7{\scriptsize $\pm$ 0.1}  & 14.6{\scriptsize $\pm$ 0.8}  & 20.8{\scriptsize $\pm$ 0.6}  & 31.0{\scriptsize $\pm$ 0.9}  &  28.1{\scriptsize $\pm$ 0.2}  & 33.2{\scriptsize $\pm$ 0.5} & 33.1{\scriptsize $\pm$ 0.1} & 31.5{\scriptsize $\pm$ 0.6} \\
    \cmidrule{2-9}
    {\bf All $\ell_\infty$ attacks} & 29.6{\scriptsize $\pm$ 0.1} & 10.5{\scriptsize $\pm$ 0.7} & 5.2{\scriptsize $\pm$ 0.6}  & 28.7{\scriptsize $\pm$ 0.9}  & 23.7{\scriptsize $\pm$ 0.2}  & 29.8{\scriptsize $\pm$ 0.1} & 29.4{\scriptsize $\pm$ 0.3} & 28.1{\scriptsize $\pm$ 0.7} \\
    \midrule
    PGD-$\ell_1$ & 38.7{\scriptsize $\pm$ 0.6}  & 44.6{\scriptsize $\pm$ 0.1}  & 44.9{\scriptsize $\pm$ 1.1}  & 36.9{\scriptsize $\pm$ 0.5}  & 44.3{\scriptsize $\pm$ 0.1} & 39.9{\scriptsize $\pm$ 0.4} & 39.2{\scriptsize $\pm$ 0.1} & 45.1{\scriptsize $\pm$ 0.5}   \\
    PGD-Foolbox  & 40.0{\scriptsize $\pm$ 0.8}  & 44.8{\scriptsize $\pm$ 0.2} & 45.2{\scriptsize $\pm$ 0.2}  & 37.6{\scriptsize $\pm$ 0.9}  & 44.7{\scriptsize $\pm$ 1.5}  & 40.6{\scriptsize $\pm$ 0.1} & 42.1{\scriptsize $\pm$ 0.9} & 45.0{\scriptsize $\pm$ 0.2}  \\
    EAD & 52.3{\scriptsize $\pm$ 1.5} & 56.3{\scriptsize $\pm$ 0.6}  & 57.3{\scriptsize $\pm$ 0.0}  & 46.7{\scriptsize $\pm$ 0.9}  & 54.6{\scriptsize $\pm$ 0.9} & 51.2{\scriptsize $\pm$ 0.2} & 52.9{\scriptsize $\pm$ 0.1} & 52.7{\scriptsize $\pm$ 0.3}  \\
    SAPA     &  46.5{\scriptsize $\pm$ 0.9}  & 52.9{\scriptsize $\pm$ 0.7}  & 53.5{\scriptsize $\pm$ 1.2}  & 40.8{\scriptsize $\pm$ 0.1}  & 50.3{\scriptsize $\pm$ 1.1}  & 46.6{\scriptsize $\pm$ 0.1} & 46.3{\scriptsize $\pm$ 0.1} & 49.3{\scriptsize $\pm$ 0.4}  \\
    \cmidrule{2-9}
    {\bf All $\ell_1$ attacks}  &  38.2{\scriptsize $\pm$ 0.7}  & 44.6{\scriptsize $\pm$ 0.1}  & 44.1{\scriptsize $\pm$ 0.4}  & 33.2{\scriptsize $\pm$ 0.2}  & 43.3{\scriptsize $\pm$ 0.2}  & 39.5{\scriptsize $\pm$ 0.4} & 39.2{\scriptsize $\pm$ 0.1} & 45.1{\scriptsize $\pm$ 0.5}  \\
    \midrule
    PGD-$\ell_2$ & 48.5{\scriptsize $\pm$ 1.1}  & 49.1{\scriptsize $\pm$ 0.1}  & 51.8{\scriptsize $\pm$ 1.8} & 42.6{\scriptsize $\pm$ 0.7} & 49.9{\scriptsize $\pm$ 1.7}  & 47.0{\scriptsize $\pm$ 0.3} & 47.4{\scriptsize $\pm$ 0.1} & 49.1{\scriptsize $\pm$ 0.4}  \\
    PGD-Foolbox  &  45.6{\scriptsize $\pm$ 0.4}  & 45.2{\scriptsize $\pm$ 0.4}  & 47.7{\scriptsize $\pm$ 0.7}  & 41.0{\scriptsize $\pm$ 0.3}  & 47.0{\scriptsize $\pm$ 1.3} & 44.9{\scriptsize $\pm$ 0.4}  & 40.9{\scriptsize $\pm$ 0.1} & 47.0{\scriptsize $\pm$ 0.2} \\
    Gaussian Noise  &  52.5{\scriptsize $\pm$ 1.3}  & 56.1{\scriptsize $\pm$ 0.6}  & 57.6{\scriptsize $\pm$ 0.3}  & 46.4{\scriptsize $\pm$ 0.9}  & 54.4{\scriptsize $\pm$ 0.8}  & 51.1{\scriptsize $\pm$ 0.0} & 52.2{\scriptsize $\pm$ 0.8} & 52.1{\scriptsize $\pm$ 0.5}  \\
    AutoAttack & 42.5{\scriptsize $\pm$ 0.8} & 41.9{\scriptsize $\pm$ 0.0}  & 44.9{\scriptsize $\pm$ 0.6}  & 38.9{\scriptsize $\pm$ 0.8}  & 44.6{\scriptsize $\pm$ 1.3}  & 42.4{\scriptsize $\pm$ 0.9}  & 37.2{\scriptsize $\pm$ 0.1} & 44.4{\scriptsize $\pm$ 0.4} \\
    Brendel \& Bethge  & 43.7{\scriptsize $\pm$ 0.4} & 44.4{\scriptsize $\pm$ 0.1} & 46.6{\scriptsize $\pm$ 1.1} & 39.2{\scriptsize $\pm$ 0.7} & 45.1{\scriptsize $\pm$ 1.6} & 43.6{\scriptsize $\pm$ 0.4} & 38.9{\scriptsize $\pm$ 0.2} & 45.4{\scriptsize $\pm$ 0.1} \\
    CWL2    & 43.5{\scriptsize $\pm$ 1.3} & 44.8{\scriptsize $\pm$ 1.1}  & 47.5{\scriptsize $\pm$ 0.7}  & 39.5{\scriptsize $\pm$ 0.4} & 46.8{\scriptsize $\pm$ 1.9} & 43.4{\scriptsize $\pm$ 0.1} & 37.6{\scriptsize $\pm$ 0.2} & 46.0{\scriptsize $\pm$ 0.4}\\
      \cmidrule{2-9}
    {\bf All $\ell_2$ attacks}  & 42.5{\scriptsize $\pm$ 0.6}  & 41.9{\scriptsize $\pm$ 0.0}  & 44.9{\scriptsize $\pm$ 0.1} & 35.8{\scriptsize $\pm$ 0.7}  & 44.6{\scriptsize $\pm$ 0.1}  & 42.4{\scriptsize $\pm$ 1.0} & 37.2{\scriptsize $\pm$ 0.1} & 44.4{\scriptsize $\pm$ 0.1}  \\
    \midrule
    Acc$^{\rm{union}}_{\rm{adv}}$ & 19.8{\scriptsize $\pm$ 1.1}  & 10.1{\scriptsize $\pm$ 0.7}  & 5.2{\scriptsize $\pm$ 0.6} & 26.1{\scriptsize $\pm$ 0.9}  & 23.6{\scriptsize $\pm$ 0.3}  & {\bf 29.8{\scriptsize $\pm$ 0.1}} & 29.4{\scriptsize $\pm$ 0.3} & 28.1{\scriptsize $\pm$ 0.8}  \\
    Acc$^{\rm{avg}}_{\rm{adv}}$ &  36.7{\scriptsize $\pm$ 0.4}  & 32.2{\scriptsize $\pm$ 0.4}  & 31.7{\scriptsize $\pm$ 0.5} & 32.8{\scriptsize $\pm$ 0.1}  & \underline{37.2{\scriptsize $\pm$ 0.2}}  & 33.5{\scriptsize $\pm$ 0.6} & 35.3{\scriptsize $\pm$ 0.2} & {\bf 39.1{\scriptsize $\pm$ 0.6}}\\
    \bottomrule
  \end{tabular}
\end{table*}

\begin{table*}[h!]
\begin{minipage}{\linewidth}
\begin{minipage}{.45\linewidth}
\begin{figure}[H]
    \centering
    \def\svgwidth{0.75\linewidth}
\begingroup%
  \makeatletter%
  \providecommand\color[2][]{%
    \errmessage{(Inkscape) Color is used for the text in Inkscape, but the package 'color.sty' is not loaded}%
    \renewcommand\color[2][]{}%
  }%
  \providecommand\transparent[1]{%
    \errmessage{(Inkscape) Transparency is used (non-zero) for the text in Inkscape, but the package 'transparent.sty' is not loaded}%
    \renewcommand\transparent[1]{}%
  }%
  \providecommand\rotatebox[2]{#2}%
  \newcommand*\fsize{\dimexpr\f@size pt\relax}%
  \newcommand*\lineheight[1]{\fontsize{\fsize}{#1\fsize}\selectfont}%
  \ifx\svgwidth\undefined%
    \setlength{\unitlength}{23.08782148bp}%
    \ifx\svgscale\undefined%
      \relax%
    \else%
      \setlength{\unitlength}{\unitlength * \real{\svgscale}}%
    \fi%
  \else%
    \setlength{\unitlength}{\svgwidth}%
  \fi%
  \global\let\svgwidth\undefined%
  \global\let\svgscale\undefined%
  \makeatother%
  \begin{picture}(1,1.302765)%
    \lineheight{1}%
    \setlength\tabcolsep{0pt}%
    \put(0,0){\includegraphics[width=\unitlength,page=1]{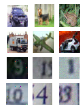}}%
    \put(0.20127478,1.30336504){\color[rgb]{0,0,0}\makebox(0,0)[lt]{\begin{minipage}{0.38663897\unitlength}\raggedright ${\ell_1}$\end{minipage}}}%
    \put(0.83026335,1.30756183){\color[rgb]{0,0,0}\makebox(0,0)[lt]{\begin{minipage}{0.36876549\unitlength}\raggedright ${\ell_\infty}$\end{minipage}}}%
    \put(0.52338322,1.30695658){\color[rgb]{0,0,0}\makebox(0,0)[lt]{\begin{minipage}{0.38663897\unitlength}\raggedright ${\ell_2}$\end{minipage}}}%
    \put(-0.00665811,0.82168963){\color[rgb]{0,0,0}\rotatebox{90}{\makebox(0,0)[lt]{\begin{minipage}{0.88340231\unitlength}\raggedright CIFAR-10\end{minipage}}}}%
    \put(-0.00908498,0.24427028){\color[rgb]{0,0,0}\rotatebox{90}{\makebox(0,0)[lt]{\begin{minipage}{0.88340231\unitlength}\raggedright SVHN\end{minipage}}}}%
  \end{picture}%
\endgroup%

    \caption{\small Visualization of the generated examples by MNG-AC along projected on $\ell_1, \ell_2$, and $\ell_\infty$-norm ball for CIFAR-10 and SVHN dataset.\label{fig:noise}}
\end{figure}
\end{minipage}
\hspace{0.2in}
\begin{minipage}{.55\linewidth}
\begin{figure}[H]
    \centering
    \def\svgwidth{\textwidth}
    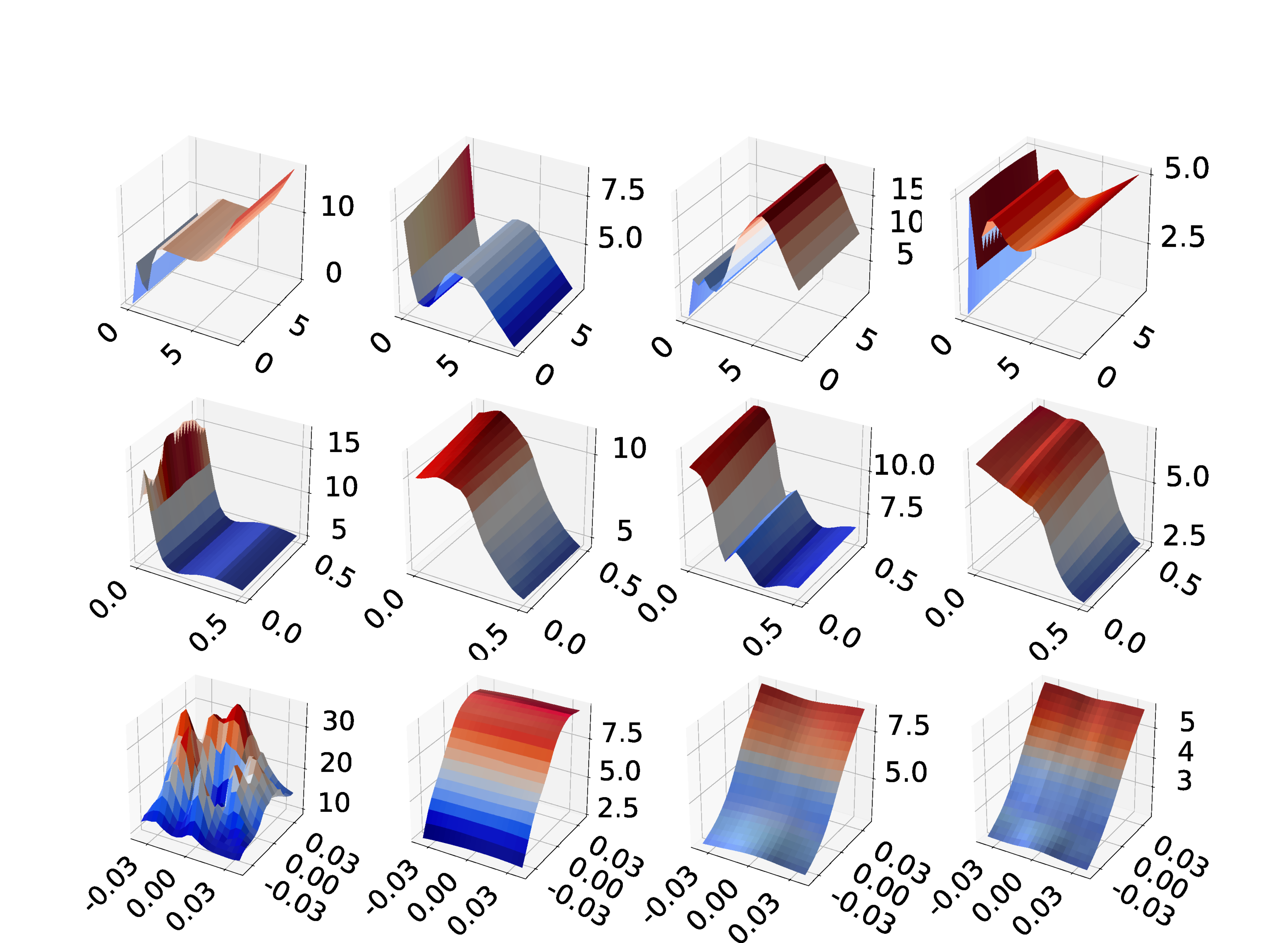

    \vspace{-0.2in}
    \caption{\small Visualization of the loss landscapes for the $\ell_1, \ell_2$, and $\ell_\infty$-norm attacks on the SVHN dataset. The rows represent the attacks and columns represent different defenses. We can observe that that MNG-AC obtains smooth loss surface across all $\ell_p$-norm attacks.
    \label{fig:appendix_loss_plots}}
\end{figure}
\end{minipage}
\end{minipage}
\end{table*}

{\bf Visuaization of loss landscape on SVHN dataset.}
\Cref{fig:appendix_loss_plots} shows the visualization of loss landscape of various methods against $\ell_\infty, \ell_1$, and $\ell_2$ norm attack for SVHN dataset on Wide ResNet 28-10 architecture. Similar to the CIFAR-10 dataset, we can observe that the loss is highly curved for multiple perturbations in the vicinity of the data point $x$ for the adversarial training trained with a single perturbation, which reflects that the gradient poorly models the global landscape. In contrast, MNG-AC achieves smoother loss surface across all types of $\ell_p$ norm attacks.

\end{document}